\documentclass[letterpaper]{article}
\usepackage{uai2019}
\usepackage[margin=1in]{geometry}
\usepackage{times}

\usepackage{ascii}
\usepackage{hyperref}       
\usepackage{url}            
\usepackage{booktabs}       
\usepackage{amsfonts, amsmath, amssymb, relsize}       
\usepackage{macros}
\usepackage{subcaption}
\usepackage{algorithmic}
\usepackage{algorithm}
\usepackage{enumitem}
\usepackage[round]{natbib}
\usepackage{graphicx}

\title{A Flexible Framework for Multi-Objective Bayesian Optimization using Random Scalarizations}

\author{
{\bf Biswajit Paria}\\
MLD, Carnegie Mellon University\\
\texttt{\asciifamily bparia@cs.cmu.edu}\\
\And
{\bf Kirthevasan Kandasamy\thanks{\ \ This work was done when KK was at CMU}}\\
EECS, UC Berkeley\\
\texttt{\asciifamily kandasamy@eecs.berkeley.edu}\\
\And
{\bf Barnab\'as P\'oczos}\\
MLD, Carnegie Mellon University\\
\texttt{\asciifamily bapoczos@cs.cmu.edu}\\
}

\begin{document}

\maketitle

\begin{abstract}
Many real world applications can be framed as multi-objective optimization problems, where we wish to simultaneously optimize for multiple criteria. Bayesian optimization techniques for the multi-objective setting are pertinent when the evaluation of the functions in question are expensive. Traditional methods for multi-objective optimization, both Bayesian and otherwise, are aimed at recovering the Pareto front of these objectives. However, in certain cases a practitioner might desire to identify Pareto optimal points only in a subset of the Pareto front due to external considerations. In this work, we propose a strategy based on random scalarizations of the objectives that addresses this problem. Our approach is able to flexibly sample from desired regions of the Pareto front and, computationally, is considerably cheaper than most approaches for MOO. We also study a notion of regret in the multi-objective setting and show that our strategy achieves sublinear regret. We experiment with both synthetic and real-life problems, and demonstrate superior performance of our proposed algorithm in terms of flexibility, scalability and regret.
\end{abstract}

\section{Introduction}
\label{sec:intro}

Bayesian optimization (BO) is a popular recipe for optimizing expensive black-box functions where the goal is to find a global maximizer of the function. Bayesian optimization has been used for a variety of practical optimization tasks such as hyperparameter tuning for machine learning algorithms, experiment design, online advertising, and scientific discovery \citep{snoek2012practical, hernandez2017parallel, martinez2007active, gonzalez2015bayesian, kandasamy2017query}.

In many practical applications however, we are required to optimize multiple objectives, and moreover, these objectives tend to be competing in nature. For instance, consider drug discovery, where each evaluation of the functions is an in-vitro experiment and as the output of the experiment, we measure the solubility, toxicity and potency of a candidate example. A chemist wishes to find a molecule that has high solubility and potency, but low toxicity. This is an archetypal example for Bayesian optimization as the lab experiment is expensive. Further, drugs that are very potent are also likely to be toxic, so these two objectives are typically competing.
Other problems include creating fast but accurate neural networks. While smaller neural networks are faster to evaluate, they suffer in terms of accuracy.

Due to their conflicting nature, all the objectives cannot be optimized simultaneously. As a result, most multi-objective optimization (MOO) approaches aim to recover the Pareto front, defined as the set of Pareto optimal points. A point is Pareto optimal if it cannot be improved in any of the objectives without degrading some other objective. More formally, given $K$ objectives $f(\xv) = (f_1(\xv), \dots, f_K(\xv)) : \Xc \to \Rb^K$ over a compact domain $\Xc \subset \Rb^d$, a point $\xv_1 \in \Xc$ is Pareto dominated by another point $\xv_2 \in \Xc$ iff $ f_k(\xv_1) \le f_k(\xv_2)\ \forall k \in [K]$ and $\exists k \in [K] \st f_k(\xv_1) < f_k(\xv_2)$, where we use the notation $[K]$ throughout the paper to denote the set $\{1, \dots, K\}$. We denote this by $f(\xv_1) \prec f(\xv_2)$. A point is Pareto optimal if it is not Pareto dominated by any other point. We use $\Xc^\star_f$ to denote the Pareto front for a multi-objective function $f$, and $f(\Xc^\star_f)$ to denote the set of Pareto optimal values, where $f(\Xv) = \{f(\xv)\ |\ \xv \in \Xv\}$ for any $\Xv \subseteq \Xc$. The traditional goal in the MOO optimization regime is to approximate the set of Pareto optimal points \citep{hernandez2016predictive, knowles2006parego, ponweiser2008multiobjective, zuluaga2013active}. 

However, in certain scenarios, it is preferable to explore only a part of the Pareto front. For example, consider the drug discovery application described above. A method which aims to find the Pareto front, might also invest its budget to discover drugs that are potent, but too toxic to administer to a human. Such scenarios arise commonly in many practical applications. Therefore, we need flexible methods for MOO that can steer the sampling strategy towards regions of the Pareto front that a domain expert may be interested in. Towards this end, we propose a Bayesian approach based on random-scalarizations in which the practitioner encodes their preferences as a prior on a set of scalarization functions.

A common approach to multi-objective optimization is to use \emph{scalarization functions}\footnote{Also known as utility functions in decision theory literature. To avoid confusion with acquisition functions which are sometimes referred to as utility functions in BO, we use the term scalarization function.} $\scale_{\lambdav}(\yv): \Rb^K \to \Rb$ \citep{roijers2013survey}, parameterized by $\lambdav$ belonging to a set $\Lambda$, and $\yv \in \Rb^K$ denoting $K$-dimensional objective values. Scalarizations are often used to convert multi-objective values to scalars, and standard Bayesian optimization methods for scalar functions are applied. Since our goal is to sample points from the Pareto front, we need additional assumptions to ensure that the utility functions are maximized for $\yv \in f(\Xc^\star_f)$. Following \citet{roijers2013survey} and \citet{zintgraf2015quality} we assume that $\scale_{\lambdav}(\yv)$ are monotonically increasing in all coordinates.  Optimizing for a single fixed scalarization amounts to the following maximization problem, which returns a single optimal point lying on the Pareto front.
\begin{equation}
    \xv^\star_{\lambdav} = \argmax_{\xv \in \Xc}\ \ \scale_{\lambdav}(f(\xv)) \label{eqn:fixed_lamb}
\end{equation}
One can verify that $\xv^\star_{\lambdav} \in \Pc_f$ follows from the monotonicity of the scalarization.
In this work, we are interested in a set of points $\Xv^\star = \{\xv_i^\star\}_{i=1}^T$ of size at most $T$, spanning a specified region of the Pareto front rather than a single point. To achieve this we take a Bayesian approach and assume a prior $p(\lambdav)$ with support on $\Lambda$, which intuitively translates to a prior on the set of scalarizations $\mathcal{S}_\Lambda = \{\scale_{\lambdav}\ |\ \lambdav \in \Lambda\}$. Thus, in place of optimizing a single scalarization, we aim to optimize over a set of scalarizations weighted by the prior $p(\lambdav)$. Each $\lambdav \in \Lambda$ maps to a pareto optimal value $f(\xv^\star_{\lambdav}) \in f(\Xc^\star_f)$. Thus, the prior $p(\lambdav)$ defines a probability distribution over the set of Pareto optimal values, and hence encodes user preference, which is depicted in Figure~\ref{fig:pareto}.

In this paper, we propose to minimize a Bayes regret which incorporates user preference through the prior and scalarization specified by the user. We propose multi-objective extensions of classical BO algorithms: upper confidence bound (UCB) \citep{auer2002using}, and Thompson sampling (TS) \citep{thompson1933likelihood} to minimize our proposed regret. At each step the algorithm computes the next point to evaluate by randomly sampling a scalarization $\scale_{\lambdav}$ using the prior $p(\lambdav)$, and optimizes it to get $\xv^\star_{\lambdav}$. Our algorithm is fully amenable to changing priors in an interactive setting, and hence can also be used with other interactive strategies in the literature. The complete algorithm is presented in Algorithm~\ref{alg:moors} and discussed in detail in Section~\ref{sec:our_approach}. While random scalarizations have been previously explored by \citet{knowles2006parego} and \citet{zhang2010expensive}, our approach is different in terms of the underlying algorithm. Furthermore, we study a more general class of scalarizations and also prove regret bounds.
As we shall see, this formulation fortunately gives rise to an extremely flexible framework that is much simpler than the existing work for MOO and computationally less expensive.
Our contributions can summarized as follows:
\begin{enumerate}[leftmargin=*,nolistsep,itemsep=4pt]
	\item We propose a flexible framework for MOO using the notion of random scalarizations.
	Our algorithm is flexible enough to sample from the entire Pareto front or an arbitrary region specified by the user.
	It is also naturally capable of sampling from non-convex regions of the Pareto front. While other competing approaches can be modified to sample from such complex regions, this seamlessly fits into our framework. In contrast to the prior work on MOBO, we consider more general scalarizations that are only required to be Lipschitz and monotonic.
    \item We prove sublinear regret bounds making only assumptions of Lipschitzness and monotonicity of the scalarization function. To our knowledge the only prior work discussing theoretical guarantees for MOO algorithms is Pareto Active Learning \citep{zuluaga2013active} with sample complexity bounds.
    \item We compare our algorithm to other existing MOO approaches on synthetic and real-life tasks. We demonstrate that our algorithm achieves the said flexibility and superior performance in terms of the proposed regret, while being computationally inexpensive.
\end{enumerate}

\subsection*{Related Work}
Most multi-objective bayesian optimization approaches aim at approximating the whole Pareto front. Predictive Entropy Search (PESMO) by \citet{hernandez2016predictive} is based on reducing the posterior entropy of the Pareto front. SMSego by \citet{ponweiser2008multiobjective} uses an optimistic estimate of the function in an UCB fashion, and chooses the point with the maximum  hypervolume improvement. Pareto Active Learning (PAL) \citep{zuluaga2013active} and $\varepsilon$-PAL \citep{zuluaga2016varepsilon} are similar to SMSego, and with theoretical guarantees. \cite{campigotto2014active} introduce another active learning approach that approximates the surface of the Pareto front. Expected hypervolume improvement (EHI) \citep{emmerich2008computation} and Sequential uncertainty reduction (SUR) \citep{picheny2015multiobjective} are two similar approaches based on maximizing the expected hypervolume. Computing the expected hypervolume is an expensive process that renders EHI and SUR computationally intractable in practice when there are several objectives.

The idea of random scalarizations has been previously explored in the following works aimed at recovering the whole Pareto front: ParEGO \citep{knowles2006parego} which uses random scalarizations to explore the whole Pareto front; MOEA/D \citep{zhang2007moea}, an evolutionary computing approach to MOO; and MOEA/D-EGO \citep{zhang2010expensive}, an extension of MOEA/D using Gaussian processes that evaluates batches of points at a time instead of a single point.
At each iteration, both ParEGO and MOEA/D-EGO sample a weight vector uniformly from the $K-1$ simplex, which is used to compute a scalar objective. The next candidate point is chosen by maximizing an \emph{off-the-shelf} acquisition function over the GP fitted on the scalar objective.
Our algorithm on the other hand, maintains $K$ different GPs, one for each objective.
Furthermore, our approach \emph{necessitates} using acquisitions specially designed for the multi-objective setting for any general scalarization; more specifically, they are generalizations of single-objective acquisitions for multiple objectives (see Table~\ref{tab:acq_funcs}).
These differences with ParEGO are not merely superficial -- our approach gives rise to a theoretical regret bound, while no such bound exists for the above methods.

Another line of work involving scalarizations include utility function based approaches. \citet{roijers2013survey,zintgraf2015quality} propose scalar utility functions as an evaluation criteria. \citet{zintgraf2018ordered, roijers2018interactive, roijers2017interactive} propose interactive strategies to maximize an unknown utility. In contrast to our approach the utility in these works is assumed to be fixed.

While there has been ample work on incorporating preferences in multi-objective optimization using evolutionary techniques \citep{deb2006reference, thiele2009preference, kim2012preference, branke2005integrating, branke2008consideration}, there has been fewer on using preferences for optimization, when using surrogate functions. Surrogate functions are essential for expensive black-box optimization. PESC \citep{garrido2016predictive} is an extension of PESM allowing to specify preferences as constraints. \citet{hakanen2017using} propose an extension of ParEGO in an interactive setting, where users provide feedback on the observations by specifying constraints on the objectives in an online fashion. \citet{yang2016preference} propose another way to take preferences into account by using truncated functions. An interesting idea proposed by \citet{sato2007controlling} uses a modified notion of Pareto dominance to prevent one or more objectives from being too small. The survey by \citet{ishibuchi2008evolutionary} on evolutionary approaches to MOO can be referred for a more extensive review.

When compared to existing work for MOO, our approach enjoys the following advantages.
\begin{enumerate}[leftmargin=*,nolistsep,itemsep=4pt]
    \item \emph{Flexibility:} Our approach allows the flexibility to specify any region of the Pareto front including non-connected regions of the Pareto front, which is not an advantage enjoyed by other methods. Furthermore, the approach is flexible enough to recover the entire Pareto front when necessary. Our approach is not restricted to linear scalarization and extends to a much larger class of scalarizations.
    \item \emph{Theoretical guarantees:} Our approach seamlessly lends itself to analysis using our proposed notion of regret, and achieves sub-linear regret bounds.
    \item \emph{Computational simplicity:} The computational complexity of our approach scales linearly with the number of objectives $K$. This is in contrast to EHI and SUR, whose complexity scales exponentially with $K$. Our method is also computationally cheaper than other entropy based methods such as PESMO.
\end{enumerate}


\section{Background}
Most BO approaches make use of a probabilistic model acting as a surrogate to the unknown function. Gaussian processes (GPs) \citet{rasmussen2006gaussian} are a popular choice for their ability to model well calibrated uncertainty at unknown points. We will begin with a brief review of GPs and single objective BO.

\textbf{Gaussian Processes.}
A Gaussian process (GP) defines a prior distribution over functions defined on some input space $\Xc$. GPs are characterized by a mean function $\mu:\Xc \mapsto \Rb$ and a kernel $\kappa: \Xc\times\Xc \mapsto \Rb$. For any function $f \sim \GP(\mu, \kappa)$ and some finite set of points $\xv_1,\dots, \xv_n \in \Xc$, the function values $f(\xv_1), \dots, f(\xv_n)$ follow a multivariate Gaussian distribution with mean $\mu$ and covariance $\Sigma$ given by $\mu_i = \mu(\xv_i),\; \Sigma_{ij} = \kappa(\xv_i, \xv_j) \ \ \forall 1 \le i,j \le n$. Examples of popular kernels include the squared exponential and the Mat\'ern kernel. The mean function is often assumed to be $0$ without any loss of generality. The posterior process, given observations $\Dc = \{(\xv_i, y_i)\}_{i=1}^{t-1}$ where $y_i = f(\xv_i) + \epsilon_i \in \Rb$, $\epsilon_i \sim \Nc(\mu, \sigma^2)$, is also a GP with the mean and kernel function given by
\begin{multline}
    \label{eqn:post_mean_var}
    \mu_t(\xv) = k^T (\Sigma + \sigma^2 I)^{-1} Y,\\
    \kappa_t(\xv, \xv') = \kappa(\xv, \xv') - k^T (\Sigma + \sigma^2 I)^{-1} k'.
\end{multline}
where $Y = \sbb{y_i}_{i=1}^t$ is the vector of observed values, $\Sigma = \sbb{\kappa(\xv_i,\xv_j)}_{i,j = 1}^t$ is the Gram matrix, $k = \sbb{\kappa(\xv, \xv_i)}_{i=1}^t$, and $k' = \sbb{\kappa(\xv', \xv_i)}_{i=1}^t$. Further details on GPs can be found in \cite{rasmussen2006gaussian}.

\textbf{Bayesian Optimization.} BO procedures operate sequentially, using past observations $\{(\xv_i,y_i)\}_{i=1}^{t-1}$ to determine the next point $\xv_t$. Given $t-1$ observations Thompson Sampling (TS) \citep{thompson1933likelihood} draws a sample $g_t$ from the posterior GP. The next candidate $\xv_t$ is choosen as $\xv_t = \argmax g_t(\xv)$. Gaussian Process UCB \citep{srinivas2010gaussian} constructs an upper confidence bound $U_t$ as $U_t(\xv) = \mu_{t-1}(\xv) + \sqrt{\beta_t} \sigma_{t-1}(\xv)$.
Here $\mu_{t-1}$ and $\sigma_{t-1}$ are the posterior mean and covariances according to equations~\ref{eqn:post_mean_var}. $\beta_t$ is a function of $t$ and the dimensionality of the input domain $\Xc$. GP-UCB stipulates that we choose $\xv_t = \argmax_{\xv\in \Xc} U_t(\xv)$. 

In this paper, we assume that the $K$ objectives $f_1, \dots, f_K$ are sampled from known GP priors $\Gc\Pc(0, \kappa_k),\ (k \in [K])$, with a common compact domain $\Xc \subset \Rb^d$.
Without loss of generality, we assume $\Xc \subseteq [0, 1]^d$. The feasible region is defined as $f(\Xc)$. We further assume that the observations are noisy, that is, $y_k = f_k(\xv) + \varepsilon_k$, where $\varepsilon_k \sim \Nc(0, \sigma_k^2),\ \forall k \in [K]$.

\section{Our Approach}
\label{sec:our_approach}

We first provide a formal description of random scalarizations, then we formulate a regret minimization problem, and finally propose multi-objective extensions of the classical UCB and TS algorithms to optimize it.

\subsection{Random Scalarizations}

\begin{figure}
    \centering
    \includegraphics[width=0.25\textwidth,trim={0.7cm 0.7cm 0 0},clip]{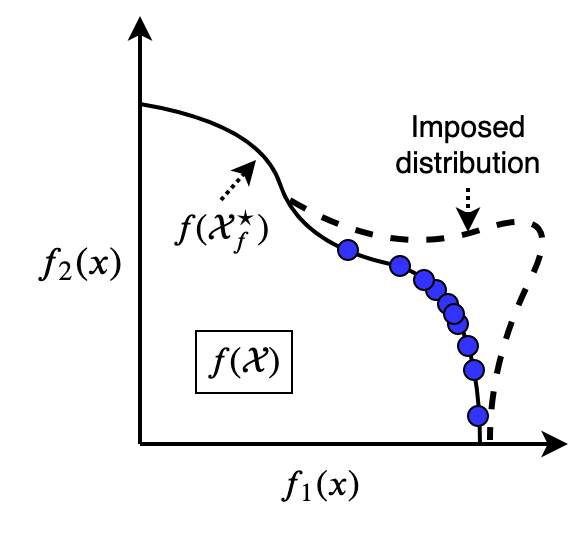}
    \caption{A prior $p(\lambdav)$ imposes a distribution on the set of Pareto optimal values. The imposed probability density is illustrated using the dotted lines. The imposed distribution leads to a concentration of the sampled values (blue circles) in the high probability region.}
    \label{fig:pareto}
\end{figure}

As discussed earlier in Section~\ref{sec:intro} in this paper we consider a set of scalarizations $\scale_{\lambdav}$ parameterized by $\lambdav \in \Lambda$. We assume a prior $p(\lambdav)$ with support $\Lambda$. We further assumed that, for all $\lambdav \in \Lambda$, $\scale_{\lambdav}$ is $L_{\lambdav}$-Lipschitz in the $\ell_1$-norm and monotonically increasing in all the coordinates. More formally,
\begin{multline}
    \label{eqn:mon_incr}
    \scale_{\lambdav}(\yv_1) - \scale_{\lambdav}(\yv_2) \le L_{\lambdav}\|\yv_1 - \yv_2\|_1,\\
    \forall \lambdav \in \Lambda,\ \ \yv_1, \yv_2 \in \Rb^d,\\
    \text{and,}\quad \scale_{\lambdav}(\yv_1) < \scale_{\lambdav}(\yv_2) \text{ whenever } \yv_1 \prec \yv_2.
\end{multline}
The Lipschitz condition can also be generalized to $\ell_p$-norms using the fact that $\|\yv\|_1 \ge K^{1-\frac{1}{p}} \|\yv\|_p$ for any $p \in [1, \infty]$ and $\yv \in \Rb^K$. Monotonicity ensures that
$$\xv^\star_{\lambdav} = \argmax_{\xv \in \Xc}\ \ \scale_{\lambdav}(f(\xv)) \in \Xc^\star_f,$$
since otherwise, if $f(\xv^\star_{\lambdav}) \prec f(\xv)$ for some $\xv \neq \xv^\star_{\lambdav}$, then we have $\scale_{\lambdav}(f(\xv^\star_{\lambdav})) < \scale_{\lambdav}(f(\xv))$, leading to a contradiction. Each $\lambdav \in \Lambda$ maps to an $\xv^\star_{\lambdav} \in \Xc_f^\star$ and a $\yv^\star =f(\xv^\star_{\lambdav}) \in f(\Xc_f^\star)$. Assuming the required measure theoretic regularity conditions hold, the prior $p(\lambdav)$ imposes a probability distribution on $f(\Xc_f^\star)$ through the above mapping as depicted in Figure~\ref{fig:pareto}.

\subsection{Bayes Regret}
\begin{figure}
    \centering
    \includegraphics[width=0.47\textwidth,trim={0.7cm 1.2cm 0 0},clip]{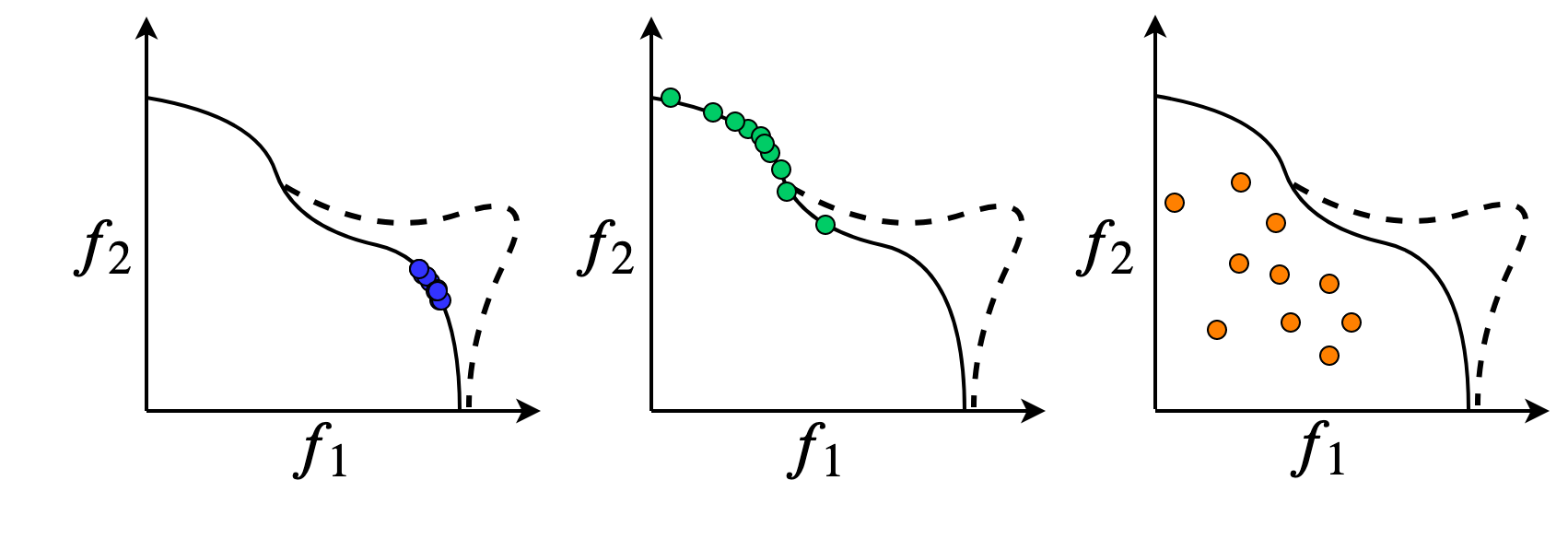}
    \caption{Three scenarios which incur a high regret: (1) The points are clustered in a small region. (2) The points are not from the desired distribution. (3) The points are not Pareto optimal.}
    \label{fig:low_regret}
\end{figure}

In contrast to (\ref{eqn:fixed_lamb}), which returns a single optimal point, in this work, we aim to return a set of points from the user specified region. Our goal is to compute a subset $\Xv \subset \Xc$ such that $f(\Xv)$ spans the high probability region of $f(\Xc^\star_f)$. This can be achieved by minimizing the following \emph{Bayes regret} denoted by $\Rc_B$,
\begin{multline}
    \label{eqn:bayes_regret}
    \!\!\!\!\!\!\Rc_B(\Xv) = \Eb_{\lambdav \sim p(\lambdav)}
    \Big(\underbrace{\max_{\xv\in \Xc}\scale_{\lambdav}(f(\xv)) - \max_{\xv \in \Xv} \scale_{\lambdav}(f(\xv))}_{\text{Pointwise regret}}\Big),\\
    \Xv^\star = \argmin_{\Xv \subset \Xc, |\Xv| \le T}\ \ \Rc_B(\Xv)\qquad\quad
\end{multline}
We now elaborate on the above expression. The pointwise regret $\max_{\xv\in \Xc}\scale_{\lambdav}(f(\xv)) - \max_{\xv \in \Xv} \scale_{\lambdav}(f(\xv))$ quantifies the regret for a particular $\lambdav$ and is analogous to the simple regret in the standard bandit setting \citep{bubeck2012regret}. $\Rc_B(T)$ similarly corresponds to the Bayes simple regret in a bandit setting.
The pointwise is minimized when $\xv^\star_{\lambdav} = \argmax_{\xv \in \Xc}\ \ \scale_{\lambdav}(f(\xv))$ belongs to $\Xv$. Since $\Xv$ is finite, the minimum may not be achieved for all $\lambdav$, as the set of optimial points can be potentially infinite. However, the regret can be small when $\exists \xv \in \Xv$ such that $f(\xv),\ f(\xv^\star_{\lambdav})$ are close, from which it follows using the Lipschitz assumption that $\scale_{\lambdav}(f(\xv^\star_{\lambdav})) - \scale_{\lambdav}(f(\xv))$ is small. Therefore, roughly speaking, the Bayes regret is minimized when the points in $\Xv$ are Pareto optimal and $f(\Xv)$ well approximates the high probability regions of $f(\Xc_f^\star)$. In this case, $\scale_{\lambdav}(f(\xv^\star_{\lambdav})) - \scale_{\lambdav}(f(\xv))$ is small for $\lambdav$s with high probabilities. Even though the rest of the regions are not well approximated, it does not affect the Bayes regret since those regions do not dominate the expectation by virtue of their low probability. This is what was desired from the beginning, that is, to compute a set of points with the majority of them spanning the desired region of interest. This is also illustrated in Figure~\ref{fig:low_regret} showing three scenarios which can incur a high regret.

It is interesting to ask, why cannot one simply maximize $$\max_{\xv \in \Xc} \Eb_{\lambdav \sim p(\lambdav)} \sbb{\scale_{\lambdav}(f(\xv))}.$$
The above expression can be maximized using a single point $\xv$ which is not the purpose of our approach. On the other hand, our proposed Bayes regret is not minimized by a single point or multiple points clustered in a small region of the Pareto front. Minimizing the pointwise regret for a single $\lambdav$ does not minimize the Bayes regret, as illustrated in Figure~\ref{fig:low_regret}. Our proposed regret has some resemblance to the expected utility metric in \citet{zintgraf2015quality}. However, the authors present it as an evaluation criteria, whereas we propose an optimization algorithm for minimizing it and also prove regret bounds on it.

\subsection{Scalarized Thompson Sampling  and UCB}

In this section we introduce Thompson Sampling and UCB based algorithms for minimizing the Bayes regret. In contrast to other methods based on random scalarizations \citep{knowles2006parego, nakayama2009sequential}, our algorithm does not convert each observation to a scalar value and fit a GP on them, but instead models them separately by maintaining a GP for each objective separately. In each iteration, we first fit a GP for each objective using the previous observations. Then we sample a $\lambdav \sim p(\lambdav)$, which is used to compute a multi-objective acquisition function based on the scalarization $\scale_{\lambdav}$. The next candidate point is chosen to be the maximizer of the acquisition function. The complete algorithm is presented in Algorithm~\ref{alg:moors} and the acquisition functions are presented in Table~\ref{tab:acq_funcs}. The acquisition function for UCB is a scalarization of the individual upper bounds of each of the objectives. Similarly, the acquisition function for TS is a scalarization of posterior samples of the $K$ objectives. 

The intuition behind our approach is to choose the $\xv_t$ that minimizes the pointwise regret for the particular $\lambdav_t$ sampled in that iteration. Looking at the expression of the Bayes regret, at a high level, it seems that it can be minimized by sampling a $\lambdav$ from the prior and choosing an $\xv_t$ that minimizes the regret for the sampled $\lambdav$. We prove regret bounds for both TS and UCB in Section~\ref{sec:theory} and show that this idea is indeed true.

\textbf{Practical Considerations.} In practice, our method requires the prior and class of scalarization functions to be specified by the user. These would typically be domain dependent. In practice, a user would also interactively update their prior based on the observations, as done in \citet{hakanen2017using, roijers2017interactive, roijers2018interactive}. Our approach is fully amenable to changing the prior interactively, and changing regions of interest. In this paper we do not propose any general methods for choosing or updating the prior, as it is not possible to do so for any general class of scalarizations. The interested readers can refer to the literature on interactive methods for MOBO. However, for the sake of demonstration we propose a simple heuristic in the experimental section.

\subsection{Computational Complexity} 
At each step all algorithms incur a cost of at most $\Oc(KT^3)$, for fitting $K$ GPs, except for ParEGO, which fits a single GP at each time step with a cost of $\Oc(T^3)$. The next step of maximizing the acquisition function differs widely across the algorithms. Computing the acquisition function at each point $\xv$ costs $\Oc(T)$ for ParEGO, and $\Oc(KT)$ for our approach. The additional factor $K$ is the price one must pay when maintaining $K$ GPs.

Apart from fitting the $K$ GPs, SMSEgo requires computing the expected hypervolume gain at each point which is much more expensive than computing the acquisitions for UCB or TS. Computing the expected hypervolume improvement in EHI is expensive and grows exponentially with $K$. PESM has a cost that is linear in $K$. However the computation involves performing expensive steps of expectation-propagation and MC estimates, which results in a large constant factor.

\begin{algorithm}[tb]
\caption{MOBO using Random Scalarizations (MOBO-RS)}
\label{alg:moors}
\begin{algorithmic}
\STATE Init $\Dc^{(0)} \gets \emptyset,\ \Gc\Pc_k^{(0)} \gets \Gc\Pc(\zero, \kappa),\ \forall k \in [K]$
\FOR{$t = 1 \to T$}
\STATE Sample $\lambdav_t \sim p(\lambdav)$
\STATE $\xv_t \gets \argmax_{\xv \in \Xc}\ \mathrm{acq}(\xv, \lambdav_t)$
\STATE \hfill\COMMENT{See Table~\ref{tab:acq_funcs} for acquisition functions}
\STATE Evaluate $\yv = f(\xv_t)$
\STATE $\Dc^{(t)} = \Dc^{(t-1)} \cup \{(\xv_t, \yv)\}$
\STATE $\Gc\Pc_k^{(t)} \gets \mathrm{post}\bb{\GP_k^{(t-1)}\ |\ (\xv_t, \yv_k)},\forall k \in [K]$
\ENDFOR
\RETURN $\Dc^{(T)}$
\end{algorithmic}
\end{algorithm}

\begin{table}
\centering
\caption{Acquisition functions used in Algorithm~\ref{alg:moors}. $\mu^{(t)}(\xv),\ \sigma^{(t)}(\xv)$ are $K$ dimensional vectors denoting the posterior means and variances at $\xv$ of the $K$ objectives respectively, in step $t$. $c$ is a hyperparameter and $d$ is dimension of the input space $\Xc$. $f_k'$ is randomly sampled from the posterior of the $k$th objective function.}
\label{tab:acq_funcs}
\begin{tabular}{cc}
\toprule
 & $\mathrm{acq}(\xv, \lambdav) $\\\midrule
UCB & $\scale_{\lambdav}\bb{\mu^{(t)}(\xv) + \sqrt{\beta_t}\sigma^{(t)}(\xv)},\ \beta_t = cd\ln t$ \\\midrule
TS & $\scale_{\lambdav}(f'(\xv))$, where $f_k' \sim \GP_k^{(t)},\ k \in [K]$ \\\bottomrule
\end{tabular}
\end{table}

\section{Regret Bounds}
\label{sec:theory}
In this section we provide formal guarantees to prove upper bounds on the Bayes regret $\Rc_B$ which goes to zero as $T \to \infty$. We also show that our upper bound is able to recover regret bounds for single objectives when $K=1$.

Analogous to the notion of regret in the single-objective setting \citep{bubeck2012regret}, we first define the instantaneous and cumulative regrets for the multi-objective optimization. The \emph{instantaneous regret} incurred by our algorithm in step $t$ is defined as,
\begin{equation}
    r(\xv_t, \lambdav_t) = \max_{\xv\in \Xc} \scale_{\lambdav_t}(f(\xv)) - \scale_{\lambdav_t}(f(\xv_t)),
\end{equation}
where $\lambdav_t$ and $\xv_t$ are the same as in Algorithm~\ref{alg:moors}. The \emph{cumulative regret} till step $T$ is defined as,

\begin{equation}
    \Rc_C(T) = \sum_{t=1}^T r(\xv_t, \lambdav_t).
\end{equation}

For convenience, we do not explicitly mention the dependency of $\Rc_C(T)$ on $\{\xv_t\}_{t=1}^T$ and $\{\lambdav_t\}_{t=1}^T$. Next, we will make a slight abuse of notation here and define $\Rc_B(T)$, the Bayes regret incurred till step $T$, as $\Rc_B(\Xv_T)$ (See Eqn.~\ref{eqn:bayes_regret}), where $\Xv_T = \{\xv_t\}_{t=1}^T$.

We further define the \emph{expected Bayes regret} as $\Eb\Rc_B(T)$, where the expectation is taken over the random process $f$, noise $\varepsilon$ and any other randomness occurring in the algorithm. Similarly, we also define the \emph{expected cumulative regret} as $\Eb\Rc_C(T)$, where the expectation is taken over all the aforementioned random variables and additionally $\{\lambdav_t\}_{t=1}^T$.
We will show that the expected Bayes regret can be upper bounded by the expected cumulative regret, which can be further upper bounded using the maximum information gain.

\textbf{Maximum Information Gain.}
The \emph{maximum information gain (MIG)} captures the notion of information gained about a random process $f$ given a set of observations. For any subset $A \subset \Xc$ define $\yv_A = \{y_a = f(a) + \varepsilon_a | a\in A\}$. The reduction in uncertainty about a random process can be quantified using the notion of information gain given by $\mathbf{I}(\yv_A;f) = \mathbf{H}(\yv_A) - \mathbf{H}(\yv_A | f)$, where $\mathbf{H}$ denotes the Shannon entropy. The maximum information gain after $T$ observations is defined as
\begin{equation}
    \gamma_T = \max_{A\subset \Xc: |A| = T}  \mathrm{I}(\yv_A; f).
\end{equation}

\textbf{Regret Bounds.} We assume that $\forall k \in [K],\ t \in [T],\ \xv \in \Xc,\ f_k(\xv)$ follows a Gaussian distribution with marginal variances upper bounded by 1, and the observation noise $\varepsilon_{tk} \sim \Nc(0, \sigma_k^2)$ is drawn independently of  everything else. Assume upper bounds $L_{\lambdav} \le L,\ \sigma^2_k \le \sigma^2,\ \gamma_{Tk} \le \gamma_T$, where $\gamma_{Tk}$ is the MIG for the $k$ th objective. When $\Xc \subseteq [0, 1]^d$, the cumulative regret after $T$ observations can be bounded as follows.

\begin{theorem}
\label{thm:rc_bound}
The expected cumulative regret for MOBO-RS after $T$ observations can be upper bounded for both UCB and TS as,
\begin{equation}
\Eb\Rc_C(T) = \Oc\bb{L\sbb{ \frac{K^2Td\gamma_{T}\ln T}{\ln\bb{1+\sigma^{-2}} }}^{1/2}}.
\end{equation}
\end{theorem}
The proof follows from Theorem~\ref{thm:final} in the appendix. The bound for single-objective BO can be recovered by setting $K=1$, which matches the bound of $\Oc(\sqrt{Td\gamma_T \ln T})$ shown in \citet{russo2014learning, srinivas2010gaussian}.
The proof is build on ideas for single objective analyses for TS and UCB \citep{russo2014learning,kandasamy2018parallelised}.

Under further assumption of the space $\Lambda$ being a bounded subset of a normed linear space, and the scalarizations $\scale_{\lambdav}$ being Lipschitz in $\lambdav$, it can be shown that $\Eb\Rc_B(T) \le \frac{1}{T}\Eb\Rc_C(T) + o(1)$, which combined with Theorem~\ref{thm:rc_bound} shows that the Bayes regret converges to zero as $T \to \infty$. A complete proof can be found in Appendix~\ref{sec:br_upper_bound}.

\section{Experimental Results}
\label{sec:experiments}

We experiment with both synthetic and real world problems. We compare our methods to the other existing MOO approaches in the literature: PESM, EHI, SMSego, ParEGO, and MOEA/D-EGO. EHI being computationally expensive is not feasible for more than two objectives. Other than visually comparing the results for three or lesser objectives we also compare them in terms of the Bayes regret defined in Eqn.~\ref{eqn:bayes_regret}.

While our method is valid for any scalarization satisfying the Lipschitz and monotonicity conditions, we demonstrate the performance of our algorithm on two commonly used scalarizations, the linear and the Tchebyshev scalarizations \citep{nakayama2009sequential} defined as,
\begin{equation}
    \begin{gathered}
    \scale_{\lambdav}^{\mathrm{lin}} (\yv) = \sum_{k=1}^K \lambdav_k \yv_k,\\
    \scale_{\lambdav}^{\mathrm{tch}} (\yv) = \min_{k=1}^K \lambdav_k (\yv_k - \zv_k),
    \end{gathered}
\end{equation}
where $\zv$ is some reference point. In both cases, $\Lambda = \{\lambdav \succ \zero\ |\ \|\lambdav\|_1 = 1\}$. It can be verified that the Lipschitz constant in both cases is upper bounded by $1$.

\textbf{Choosing the weight distribution $p(\lambdav)$.} While the user has the liberty to choose any distribution best suited for the application at hand, for demonstration we show one possible way. A popular way of specifying user preferences is by using bounding boxes \citep{hakanen2017using}, where the goal is to satisfy $f_k(\xv) \in [a_k, b_k],\ \forall 1\le k \le K$. We convert bounding boxes to a weight distribution using a heuristic described below.

\begin{figure}
\centering
\includegraphics[width=0.18\textwidth, trim={0 0.6cm 0 0},clip]{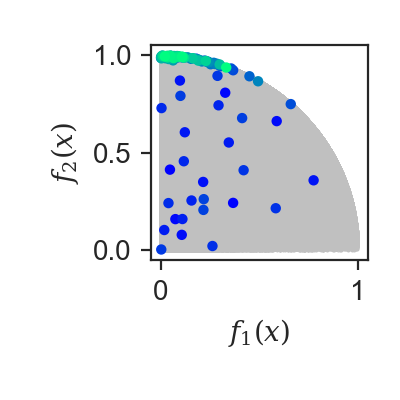}
\includegraphics[width=0.18\textwidth, trim={0 0.6cm 0 0},clip]{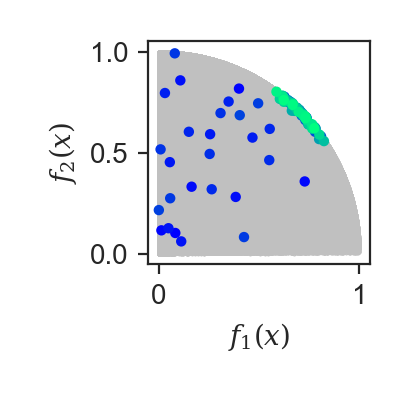}
\caption{The feasible region is shown in grey. The color of the sampled points corresponds to the iteration they were sampled in, with brighter colors being sampled in the later iterations.}
\label{fig:circle_example}
\end{figure}

\begin{figure}
    \centering
    \includegraphics[width=0.27\textwidth,trim={0.8cm 2.9cm 0 0},clip]{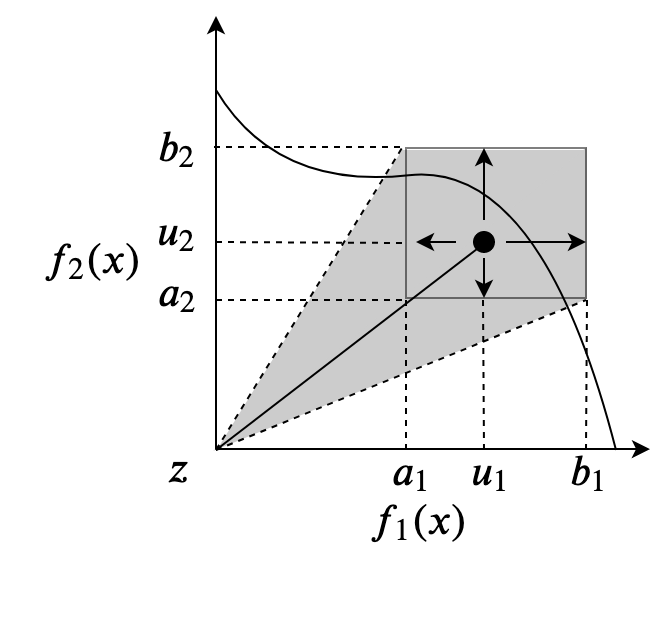}
    \caption{Weight distribution from bounding box.}
    \label{fig:box_to_dist}
\end{figure}

For the linear scalarization, it can be verified that the regret is minimized when $\yv$ is pareto optimal, and the normal vector at the surface of the Pareto front at $\yv$ has the same direction as $\lambdav$. This is illustrated using a simple example in Figure~\ref{fig:circle_example}. Consider two simple objectives $f_1(x, y) = xy, f_2(x, y) = y\sqrt{1-x^2}$. Sampling $\lambdav = \sbb{\frac{u}{u+1}, \frac{1}{u+1}}$ where $u \sim \unif{0, 0.3}$, results in the first figure. In this example we have $\lambdav_1$ smaller than $\lambdav_2$, resulting in exploration of the region where $f_2(x,y)$ is high. Whereas sampling $\lambdav = \sbb{\frac{u}{u+v}, \frac{v}{u+v}}$ where $u, v \sim \unif{0.5, 0.7}$ results in the second figure since both components of $\lambdav$ have similar magnitudes. This idea leads to the following heuristic to convert bounding boxes to a sampling strategy. We sample as $\lambdav = \uv / \|\uv\|_1$ where $\uv_k \sim \unif{a_k, b_k},\ k \in [K]$. The intuition behind this approach is shown in Figure~\ref{fig:box_to_dist}. Such a weight distribution roughly samples points from inside the bounding box.

For the Tchebychev scalarization, at the optimum, $\yv-\zv$ is inversely proportional to $\lambdav$. For the purpose of demonstration and comparison we would like both the scalarization to obtain similar objective values. Therefore, we reuse the $\lambdav$ sampled for the linear scalarization to get $\lambdav_{\mathrm{tch}} = \lambdav'/\|\lambdav'\|_1$ where $\lambdav' = (1/\lambdav_1, \dots, 1/\lambdav_K)$. We have normalized the vector so that it lies in $\Lambda$.

In order to explore the whole Pareto front, one can also specify a \emph{flat} distribution. For instance consider the Dirichlet distribution on the simplex $\{\xv\in \Rb^K\ |\ \sum_{k=1}^K \xv_k = 1, \xv \succ 0\}$. One can sample from the Dirichlet distribution as $\lambdav \sim \mathrm{Dir}(1, \dots, 1)$, which roughly provides equal weight to all the objectives leading to exploration of the whole Pareto front. Other strategies include $\lambdav = |\lambdav'|/\|\lambdav'\|_1$ where $\lambdav' \sim \Nc(\zero, \bm{I})$.

Other possible ways of choosing the weight vector includes learning the distribution of the weight vector from interactive user feedback. In fact, our framework also allows us to perform a joint posterior inference on the GP model and the weight distribution, thus learning the weight distribution in a more principled manner. While we leave these methods to future work, this demonstrates the flexibility of our framework.

\textbf{Experimental Setup.} For all our experiments, we use the squared exponential function as the GP kernel (in practice, this is a hyperparameter that must be specified by the user), given by $\kappa(\xv_1, \xv_2) = s \exp\bb{-\|\xv_1 - \xv_2\|^2/(2\sigma^2)}$, where $s$ and $\sigma$ are parameters that are estimated during optimization. We perform experiments with both TS and UCB using both kinds of scalarizations. In Eqn.~\ref{eqn:bayes_regret}, we observe that the term $\Eb_{\lambdav} \max_{\xv \in \Xc} \scale_{\lambdav}(f(\xv))$ is independent of the algorithm, hence it is sufficient to plot $-\Eb_{\lambdav} \max_{\xv \in \Xv} \scale_{\lambdav}(f(\xv))$. In all our experiments, we plot this expression, thus avoiding computing the global maximum of an unknown function. For the purposes of computing the Bayes simple regret, we linearly map the objective values to $[0, 1]$ so that the values are of reasonable magnitude. This however is not a requirement of our algorithm. Further experimental details can be found in the Appendix. The implementation can be found in Dragonfly\footnote{\url{https://github.com/dragonfly/dragonfly}}, a publicly available python library for scalable Bayesian optimization \citep{kandasamy2019tuning}.

\begin{figure}
\centering
\begin{subfigure}[t]{0.16\textwidth}
\includegraphics[width=\textwidth]{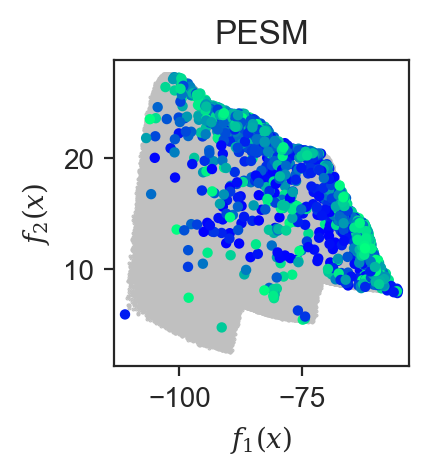}
\end{subfigure}%
\begin{subfigure}[t]{0.16\textwidth}
\includegraphics[width=\textwidth]{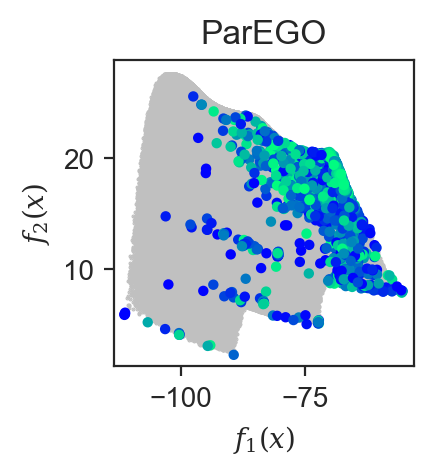}
\end{subfigure}%
\begin{subfigure}[t]{0.16\textwidth}
\includegraphics[width=\textwidth]{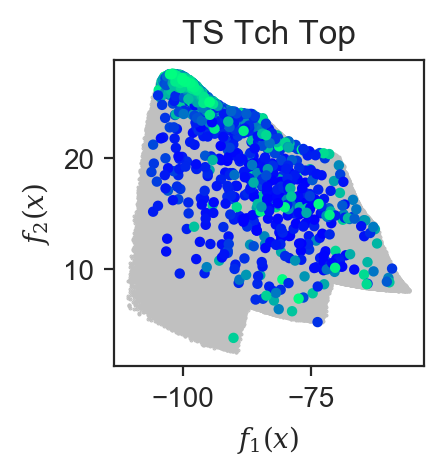}
\end{subfigure}
\begin{subfigure}[t]{0.16\textwidth}
\includegraphics[width=\textwidth]{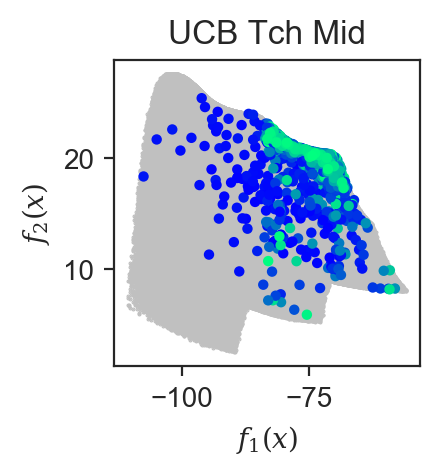}
\end{subfigure}%
\begin{subfigure}[t]{0.16\textwidth}
\includegraphics[width=\textwidth]{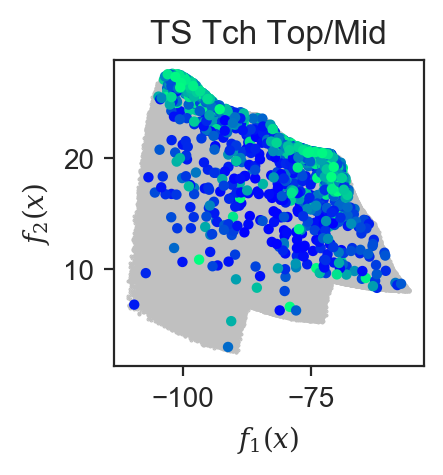}
\end{subfigure}%
\begin{subfigure}[t]{0.16\textwidth}
\includegraphics[width=\textwidth]{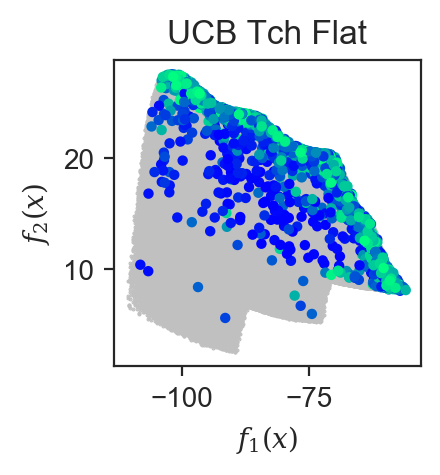}
\end{subfigure}
\caption{The feasible region is shown in grey. The color of the sampled points corresponds to the iteration they were sampled in, with brighter colors being sampled in the later iterations. The figure titles denote the method used and the region sampled. A complete set of results is presented in the Appendix.}
\label{fig:scatter_2d}
\end{figure}

\begin{figure*}
\centering
\begin{subfigure}[t]{0.30\textwidth}
\includegraphics[width=\textwidth]{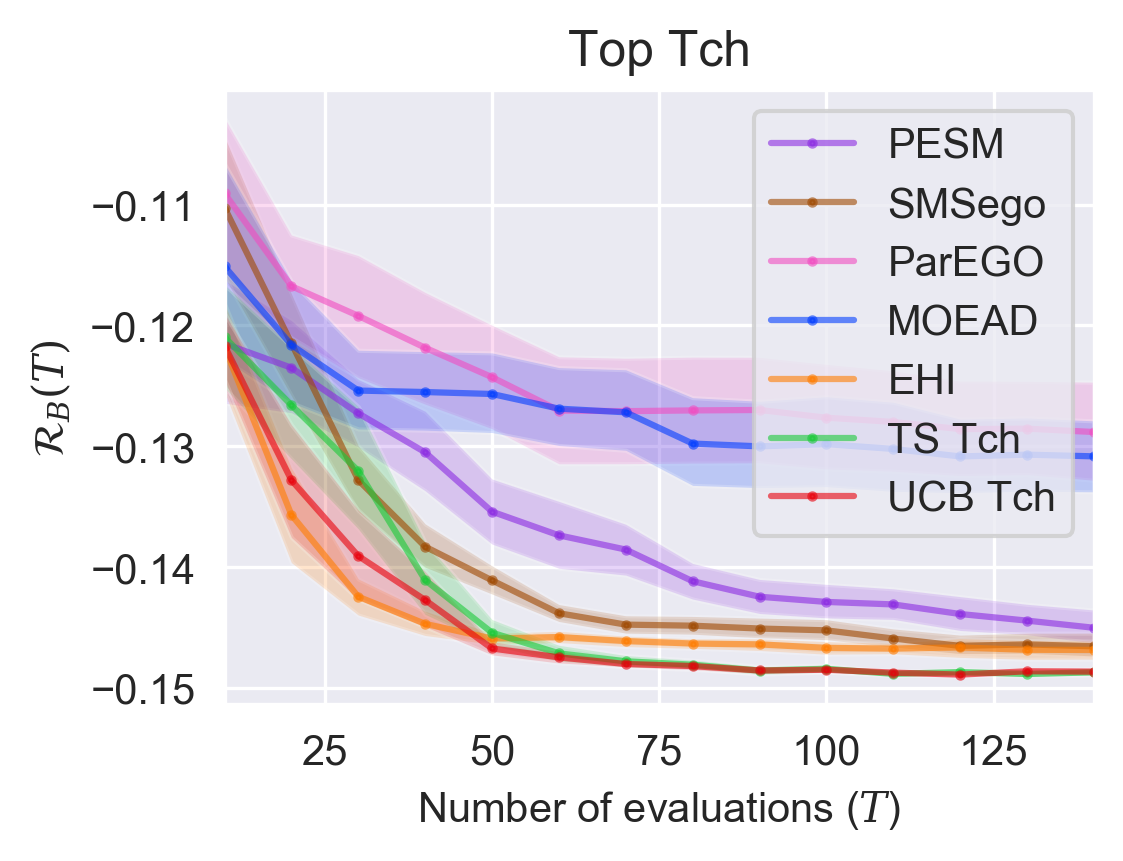}
\end{subfigure}
\begin{subfigure}[t]{0.30\textwidth}
\includegraphics[width=\textwidth]{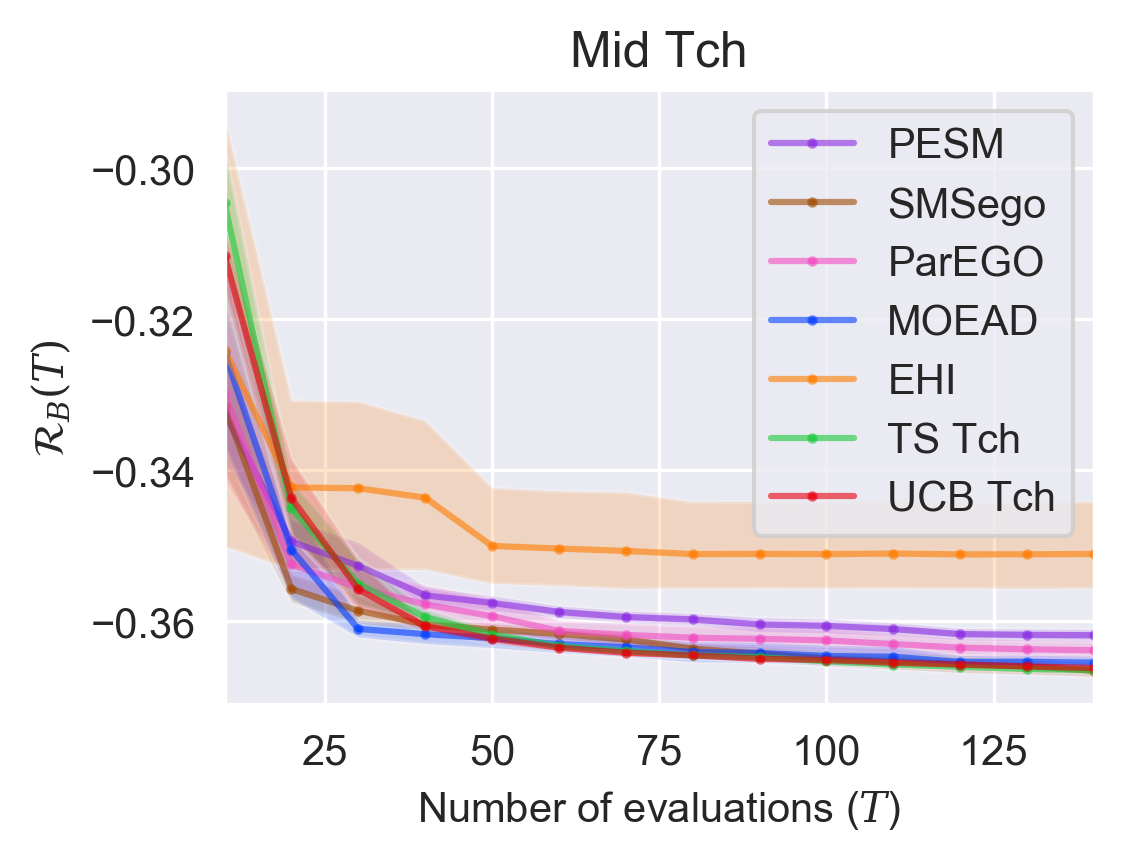}
\end{subfigure}
\begin{subfigure}[t]{0.30\textwidth}
\includegraphics[width=\textwidth]{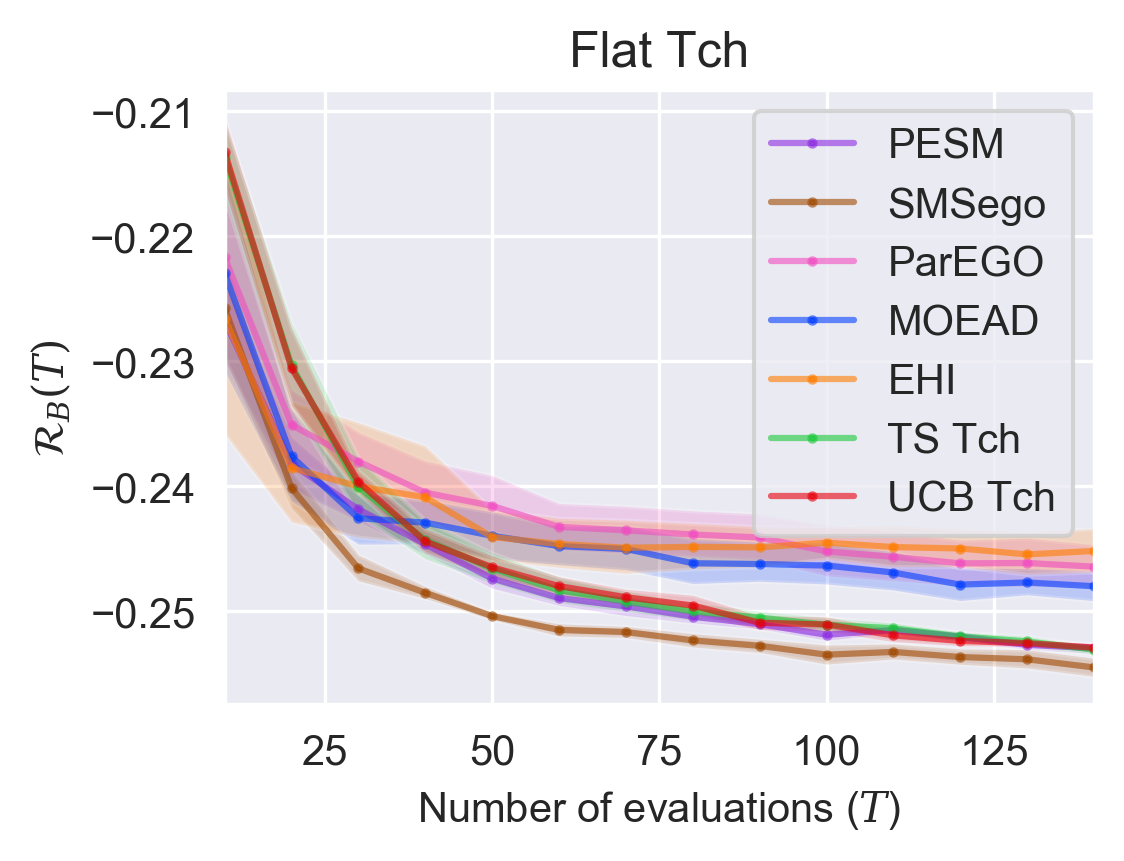}
\end{subfigure}
\caption{Bayes regret plots for the synthetic two-objective function. The mean and the 90\% confidence interval were computed over 10 independent runs. The figure titles denote the sampling region and the scalarization used. We refer the reader to the appendix for results on linear scalarization.}
\label{fig:bsr_2d_syn}
\end{figure*}

\textbf{Synthetic two-objective function.}
We construct a synthetic two-objective optimization problem using the Branin-4 and CurrinExp-4 functions as the two objectives respectively. These are the 4-dimensional counterparts of the Branin and CurrinExp functions \citep{lizotte2008practical}, each mapping $[0, 1]^4 \to \Rb$. For this experiment we specify the bounding boxes $[(a_1, b_1), (a_2, b_2)]$. We sample from three different regions, which we label as \emph{top}: $[(-110, -95), (23, 27)]$, \emph{mid}:$[(-80, -70), (16, 22)]$, and \emph{flat}: where we sample from a flat distribution. We also sample from a mixture of the \emph{top} and \emph{mid} distributions denoted by \emph{top/mid}, thus demonstrating sampling from non-connected regions in the Pareto front. Figure~\ref{fig:scatter_2d} shows a scatter plot of the sampled values for the various methods. The simple regret plots are shown in Figure~\ref{fig:bsr_2d_syn}.


\begin{figure*}
\centering
\begin{subfigure}[b]{0.30\textwidth}
    \centering
    \includegraphics[width=\textwidth]{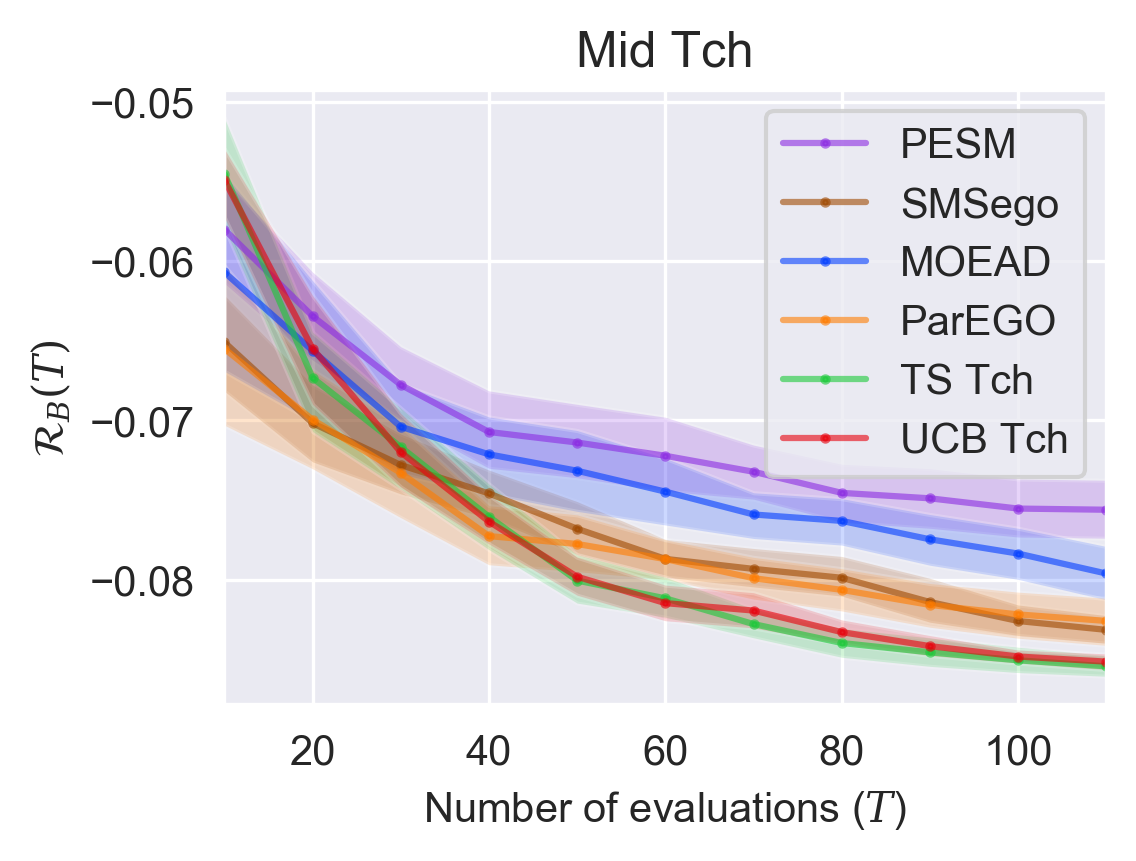}
    \subcaption{Synthetic 6x6 function}
\end{subfigure}
\begin{subfigure}[b]{0.30\textwidth}
    \centering
    \includegraphics[width=\textwidth]{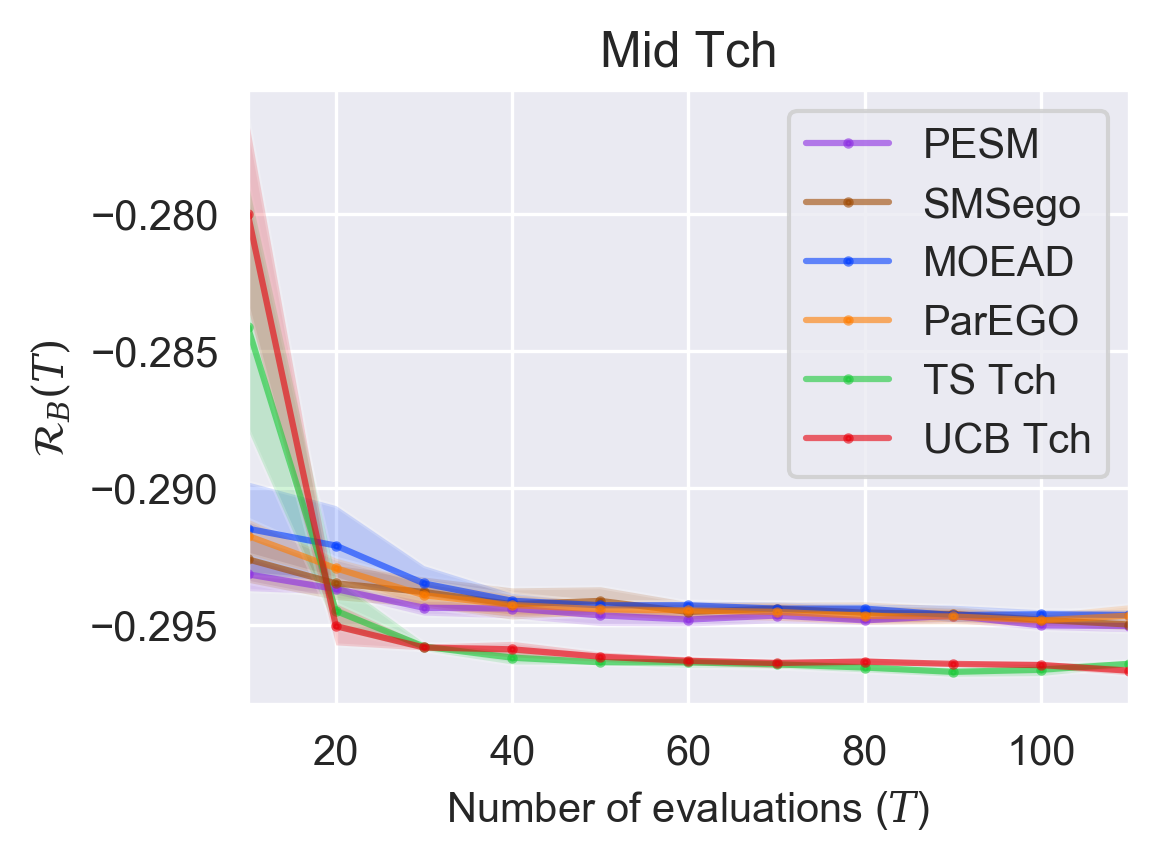}
    \subcaption{LSH Glove}
\end{subfigure}
\begin{subfigure}[b]{0.30\textwidth}
    \centering
    \includegraphics[width=\textwidth]{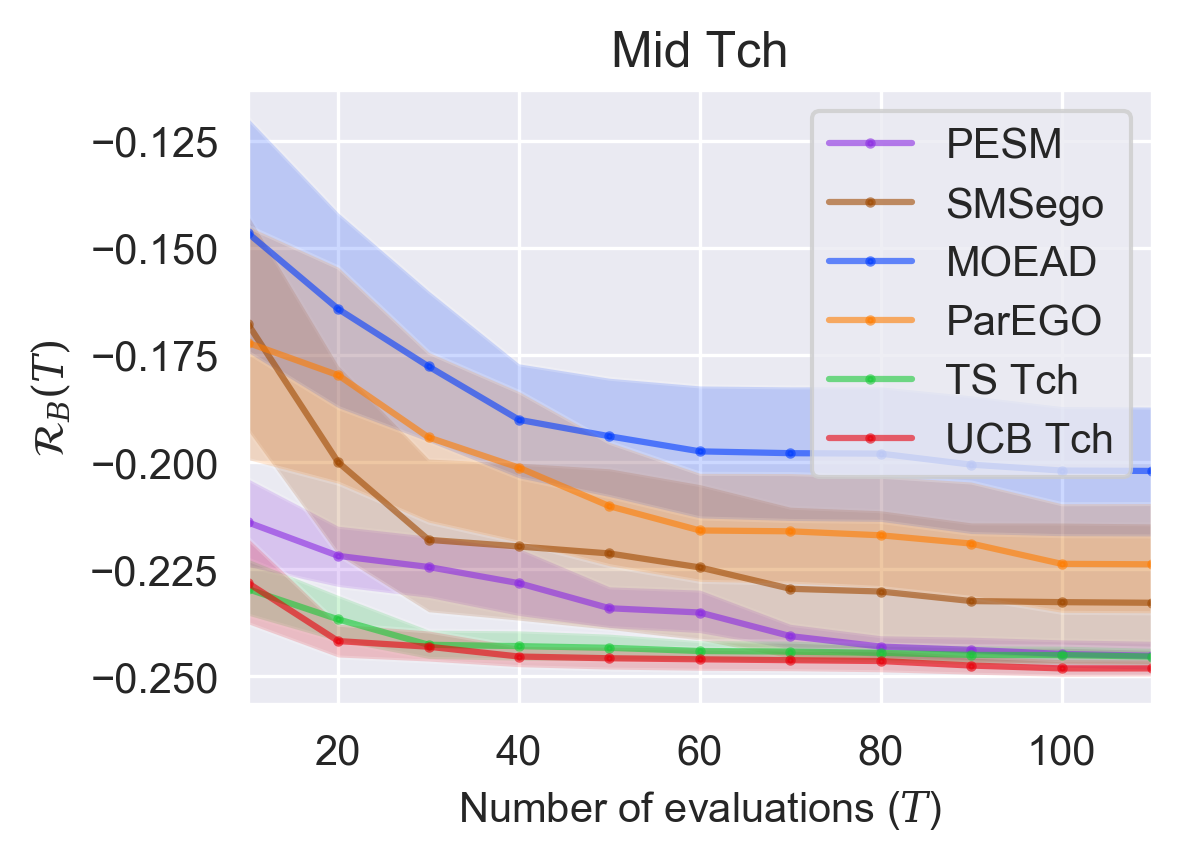}
    \subcaption{Viola Jones}
\end{subfigure}
\caption{Bayes regret plots. The mean and the 90\% confidence interval were computed over 5 runs. The figure titles denote the region sampled and the scalarization used. A complete set of plots can be found in the appendix.}
\label{fig:bsr_high_dim}
\end{figure*}

\textbf{Synthetic six-objective function.}
To show the viability of our method in high-dimensions, we sample six random functions $f_k:\Rb^6 \to \Rb,\ f_k \sim \GP(\zero, \kappa),\ k \in [6]$ where $\kappa$ is the squared exponential kernel. Devoid of any domain knowledge about this random function, we linearly transform the objectives values to $[0,1]$ for simplicity. We specify the bounding box as $[a_k, b_k] = [2/3, 1],\ \forall k \in [6]$ and denote it as the \emph{mid} region, as the weight samples are of similar magnitude. The simple regret plot for this experiment is shown in Figure~\ref{fig:bsr_high_dim}.

\textbf{Locality Sensitive Hashing.}
Locality Sensitive Hashing (LSH) \citep{andoni2015practical} is a randomized algorithm for computing the $k$-nearest neighbours. LSH involves a number of tunable parameters: the number of hash tables, number of hash bits, and the number of probes to make for each query. The parameters affect the average query time, precision and memory usage. While increasing the number of hash tables results in smaller query times, it leads to an increase in the memory footprint. Similarly, while increasing the number of probes leads to a higher precision, it increases the query time. We explore the trade-offs between these three objectives.

\begin{figure}
\centering
\begin{subfigure}[b]{0.25\textwidth}
\includegraphics[width=\textwidth]{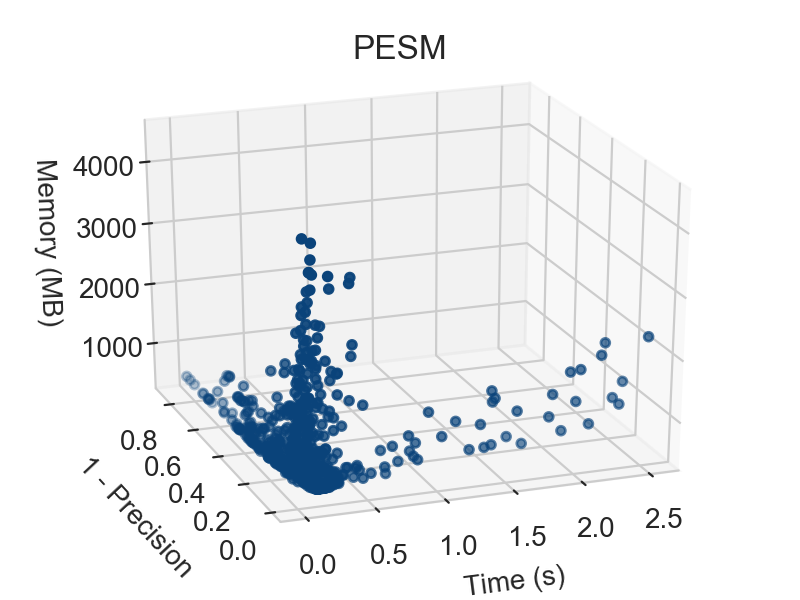}
\end{subfigure}%
\begin{subfigure}[b]{0.25\textwidth}
\includegraphics[width=\textwidth]{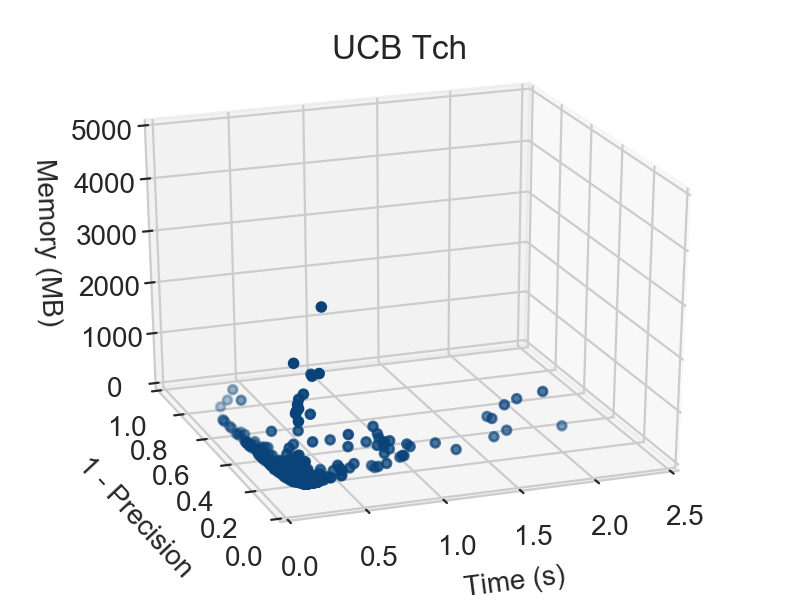}
\end{subfigure}
\caption{Sampled values for the LSH-Glove experiment over 5 independent runs. The figure titles denote the method used. A complete set of plots can be found in the appendix.}
\label{fig:lsh_samples}
\end{figure}

We run LSH using the publicly available FALCONN library\footnote{\url{https://github.com/falconn-lib/falconn}} on Glove word embeddings \citep{pennington2014glove}. We use the Glove Wikipedia-Gigaword dataset trained on $6$B tokens with a vocabulary size of $400$K and $300$-d embeddings. Given a word embedding, finding the nearest word embedding from a dictionary of word embeddings is a common task in NLP applications. We consider the following three objectives to minimize with their respective bounding boxes: Time (s) $[0.0, 0.65]$, $1-$Precision $[0.0, 0.35]$, and the Memory (MB) $[0, 1600]$. The SR plots are shown in Figure~\ref{fig:bsr_high_dim} and the sampled objective values in Figure~\ref{fig:lsh_samples}.

\textbf{Viola Jones.} The Viola Jones algorithm \citep{viola2001rapid} is a fast stagewise face detection algorithm. At each stage a simple feature detector is run over the image producing a real value. If the value is smaller than a threshold the algorithm exits with a decision, otherwise the image is processed by the next stage and so on. The Viola Jones pipeline has 27 tunable thresholds. We treat these thresholds as inputs and optimize for Sensitivity, Specificity, and the Time per query. We consider the following three objectives to minimize with their bounding boxes: $1-$Sensitivity $[0, 0.3]$, $1-$Specificity $[0, 0.13]$, and Time per query $[0, 0.07]$. Figure~\ref{fig:bsr_high_dim} shows the regret plot for this experiment.

\textbf{Results and Discussion.} Figures~\ref{fig:scatter_2d} and~\ref{fig:lsh_samples} show the sampling patterns of our proposed approach for the synthetic 2-d and the LSH glove experiment. We observe that our approach successfully samples from the specified region after some initial exploration, leading to a high concentration of points in the desired part of the Pareto front in the later iterations.

In Figures~\ref{fig:bsr_2d_syn} and~\ref{fig:bsr_high_dim} we observe that the proposed approach achieves a smaller or comparable regret compared to the other baselines. We notice that the improvement is most significant for the high dimensional experiments. A plausible explanation for this could be that learning high dimensional surfaces have a much higher sample complexity. However, our since our approach learns only a part of the Pareto front, it is able to achieve a small regret in a few number of samples, thus demonstrating the effectiveness of our approach.

\section{Conclusion}
In this paper we proposed a MOBO algorithm for efficient exploration of specific parts of the Pareto front.  We experimentally showed that our algorithm can successfully sample from a specified region of the Pareto front as is required in many applications, but is still flexible enough to sample from the whole Pareto front. Furthermore, our algorithm is computationally cheap and scales linearly with the number of objectives.

Our approach also lends itself to a notion of regret in the MO setting that also captures user preferences; with the regret being high if not sampled from the specified region or sampled outside of it. We provided a theoretical proof of the fact that our algorithm achieves a zero regret in the limit under necessary regularity assumptions. We experimentally showed that our approach leads to a smaller or comparable regret compared to the baselines.

\textbf{Acknowledgements.} This project has been been supported in part by NSF grant IIS-1563887.





\bibliography{ref}
\bibliographystyle{plainnat}





\clearpage

\appendix
\section*{\Large Appendix}
\section{Plots}
\begin{figure*}
\centering
\begin{subfigure}[t]{0.20\textwidth}
\includegraphics[width=\textwidth]{figures/Branin-4_CurrinExp-4_PESM.png}
\end{subfigure}%
\begin{subfigure}[t]{0.20\textwidth}
\includegraphics[width=\textwidth]{figures/Branin-4_CurrinExp-4_ParEGO.png}
\end{subfigure}%
\begin{subfigure}[t]{0.20\textwidth}
\includegraphics[width=\textwidth]{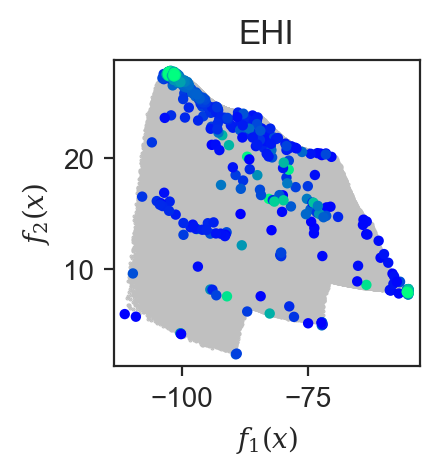}
\end{subfigure}%
\begin{subfigure}[t]{0.20\textwidth}
\includegraphics[width=\textwidth]{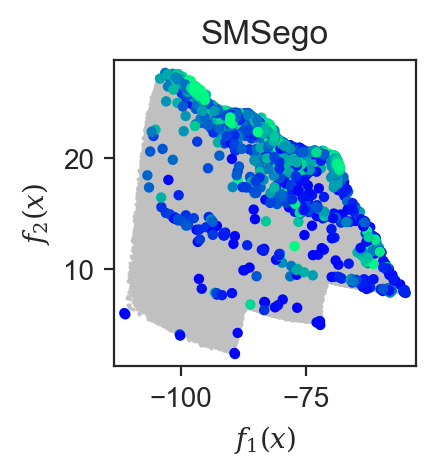}
\end{subfigure}%
\begin{subfigure}[t]{0.20\textwidth}
\includegraphics[width=\textwidth]{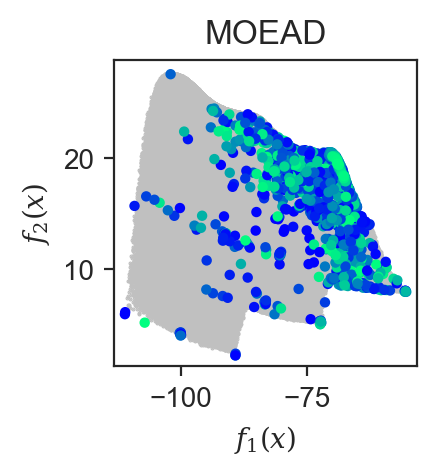}
\end{subfigure}
\begin{subfigure}[t]{0.20\textwidth}
\includegraphics[width=\textwidth]{figures/Branin-4_CurrinExp-4_ts_moc_high.png}
\end{subfigure}
\begin{subfigure}[t]{0.20\textwidth}
\includegraphics[width=\textwidth]{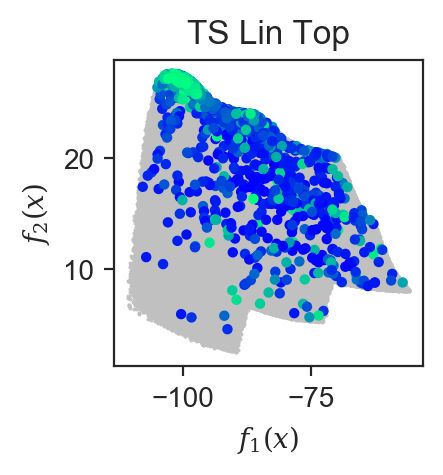}
\end{subfigure}
\begin{subfigure}[t]{0.20\textwidth}
\includegraphics[width=\textwidth]{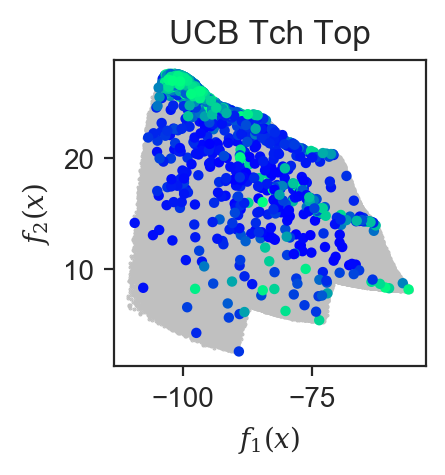}
\end{subfigure}
\begin{subfigure}[t]{0.20\textwidth}
\includegraphics[width=\textwidth]{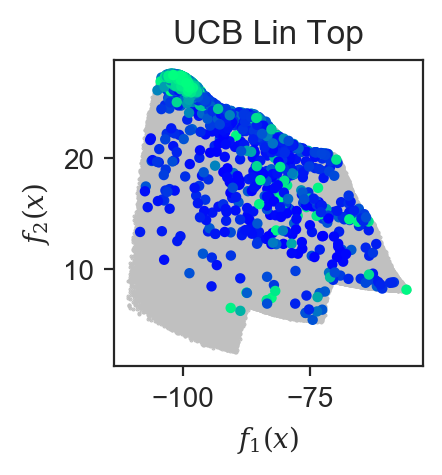}
\end{subfigure}
\begin{subfigure}[t]{0.20\textwidth}
\includegraphics[width=\textwidth]{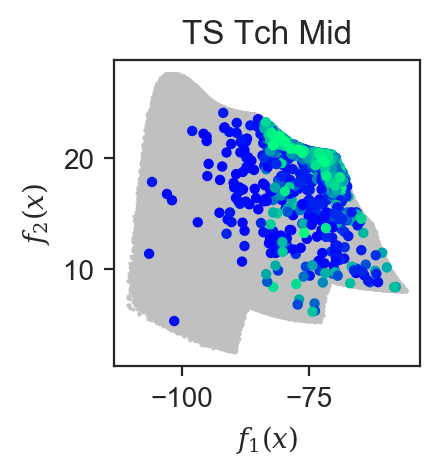}
\end{subfigure}
\begin{subfigure}[t]{0.20\textwidth}
\includegraphics[width=\textwidth]{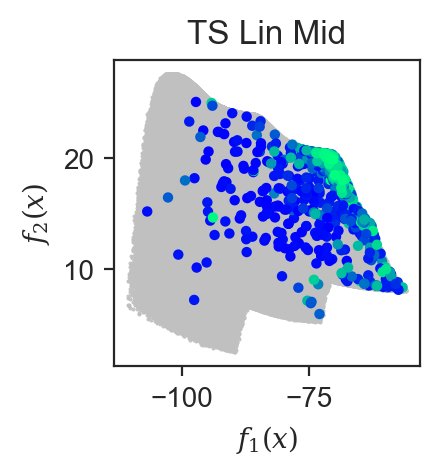}
\end{subfigure}
\begin{subfigure}[t]{0.20\textwidth}
\includegraphics[width=\textwidth]{figures/Branin-4_CurrinExp-4_ucb_moc_mid.png}
\end{subfigure}
\begin{subfigure}[t]{0.20\textwidth}
\includegraphics[width=\textwidth]{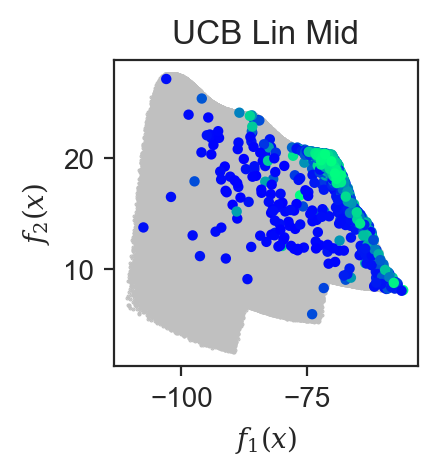}
\end{subfigure}
\begin{subfigure}[t]{0.20\textwidth}
\includegraphics[width=\textwidth]{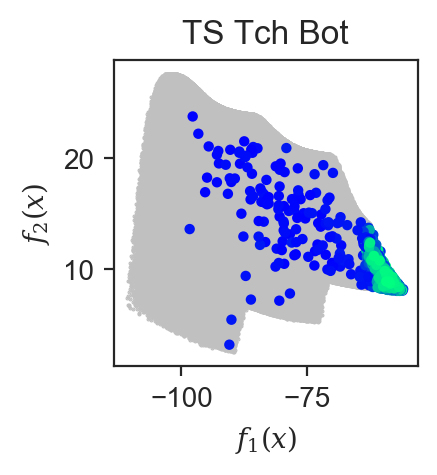}
\end{subfigure}
\begin{subfigure}[t]{0.20\textwidth}
\includegraphics[width=\textwidth]{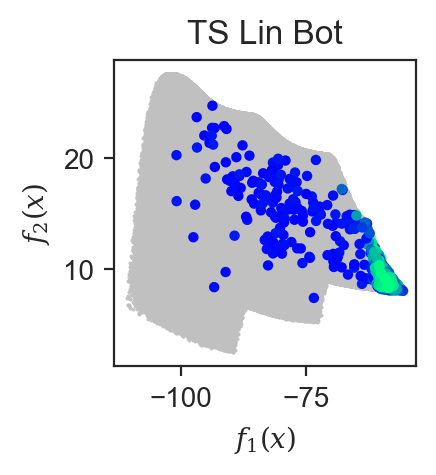}
\end{subfigure}
\begin{subfigure}[t]{0.20\textwidth}
\includegraphics[width=\textwidth]{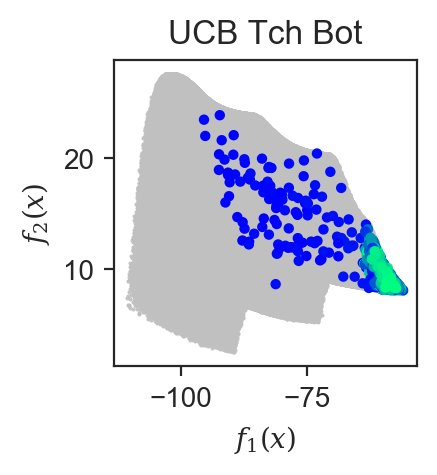}
\end{subfigure}
\begin{subfigure}[t]{0.20\textwidth}
\includegraphics[width=\textwidth]{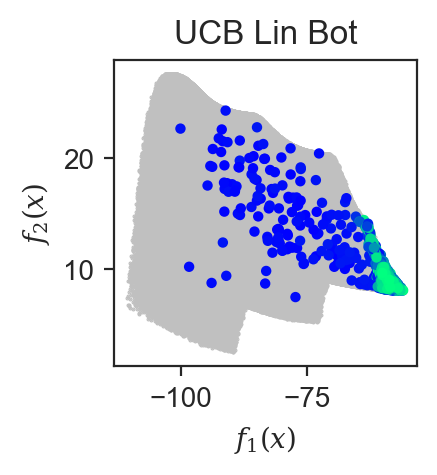}
\end{subfigure}
\begin{subfigure}[t]{0.20\textwidth}
\includegraphics[width=\textwidth]{figures/Branin-4_CurrinExp-4_ts_moc_high_mid.png}
\end{subfigure}
\begin{subfigure}[t]{0.20\textwidth}
\includegraphics[width=\textwidth]{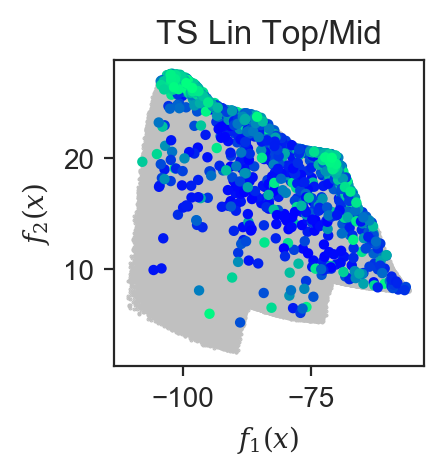}
\end{subfigure}
\begin{subfigure}[t]{0.20\textwidth}
\includegraphics[width=\textwidth]{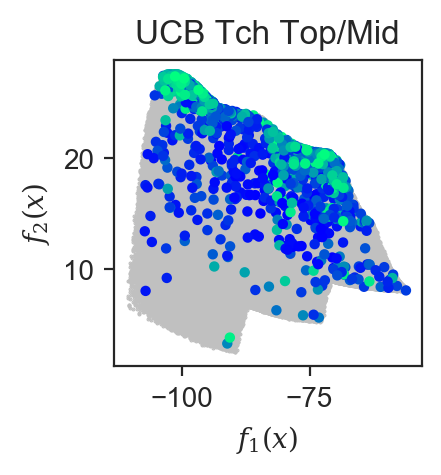}
\end{subfigure}
\begin{subfigure}[t]{0.20\textwidth}
\includegraphics[width=\textwidth]{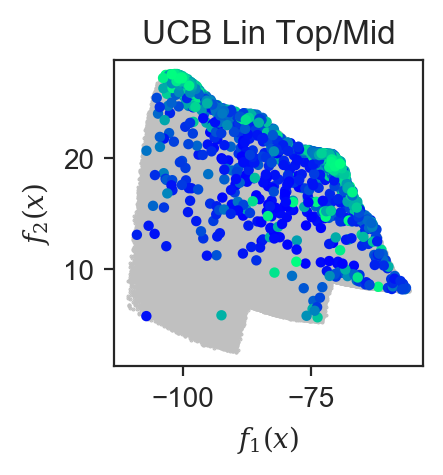}
\end{subfigure}
\caption{The plots show the sampled values for various algorithms and sampling regions. The feasible region is shown in grey. The color of the sampled points corresponds to the iteration in which they were sampled. Brighter colors were sampled in the later iterations. The figure titles denote the method used and the region sampled.}
\label{fig:all_scatter_2d}
\end{figure*}

\begin{figure*}
\centering
\begin{subfigure}[t]{0.30\textwidth}
\includegraphics[width=\textwidth]{./figures/sregret_Branin-4_CurrinExp-4_high_c.png}
\end{subfigure}
\begin{subfigure}[t]{0.30\textwidth}
\includegraphics[width=\textwidth]{./figures/sregret_Branin-4_CurrinExp-4_mid_c.png}
\end{subfigure}
\begin{subfigure}[t]{0.30\textwidth}
\includegraphics[width=\textwidth]{./figures/sregret_Branin-4_CurrinExp-4_full_c.png}
\end{subfigure}
\begin{subfigure}[t]{0.30\textwidth}
\includegraphics[width=\textwidth]{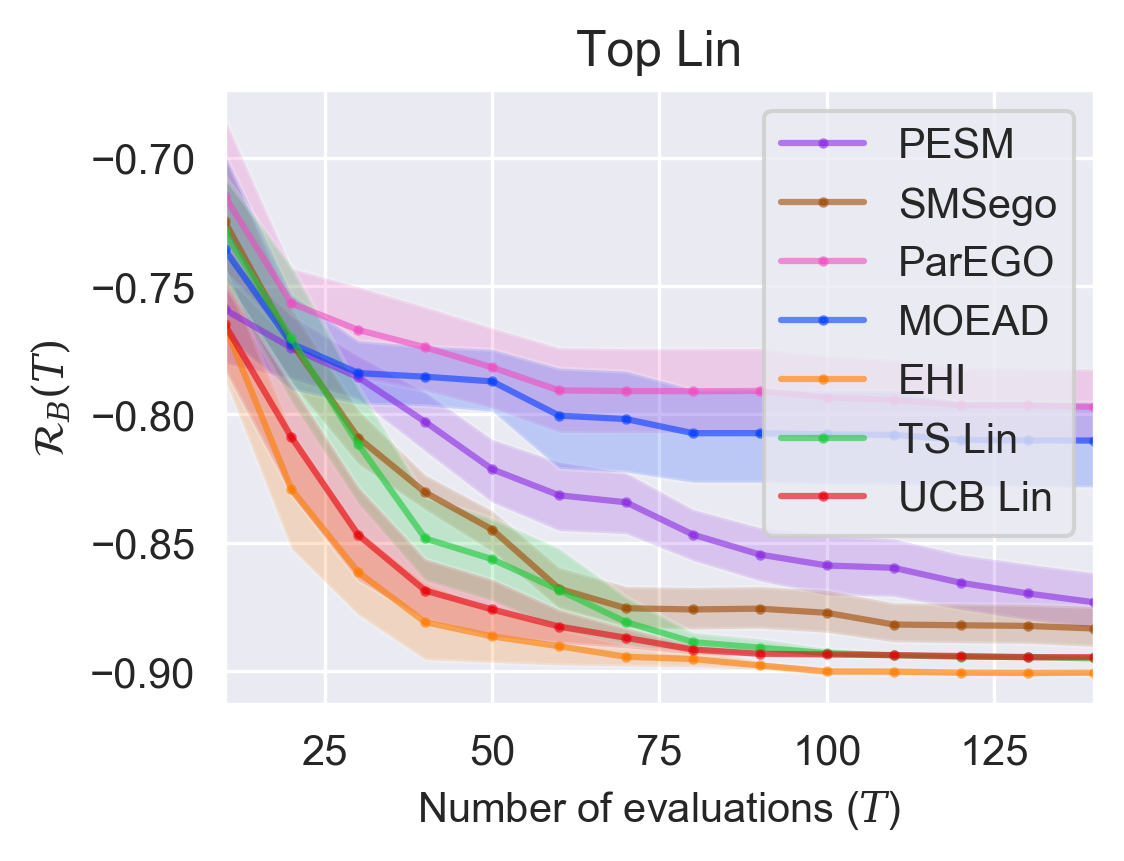}
\end{subfigure}
\begin{subfigure}[t]{0.30\textwidth}
\includegraphics[width=\textwidth]{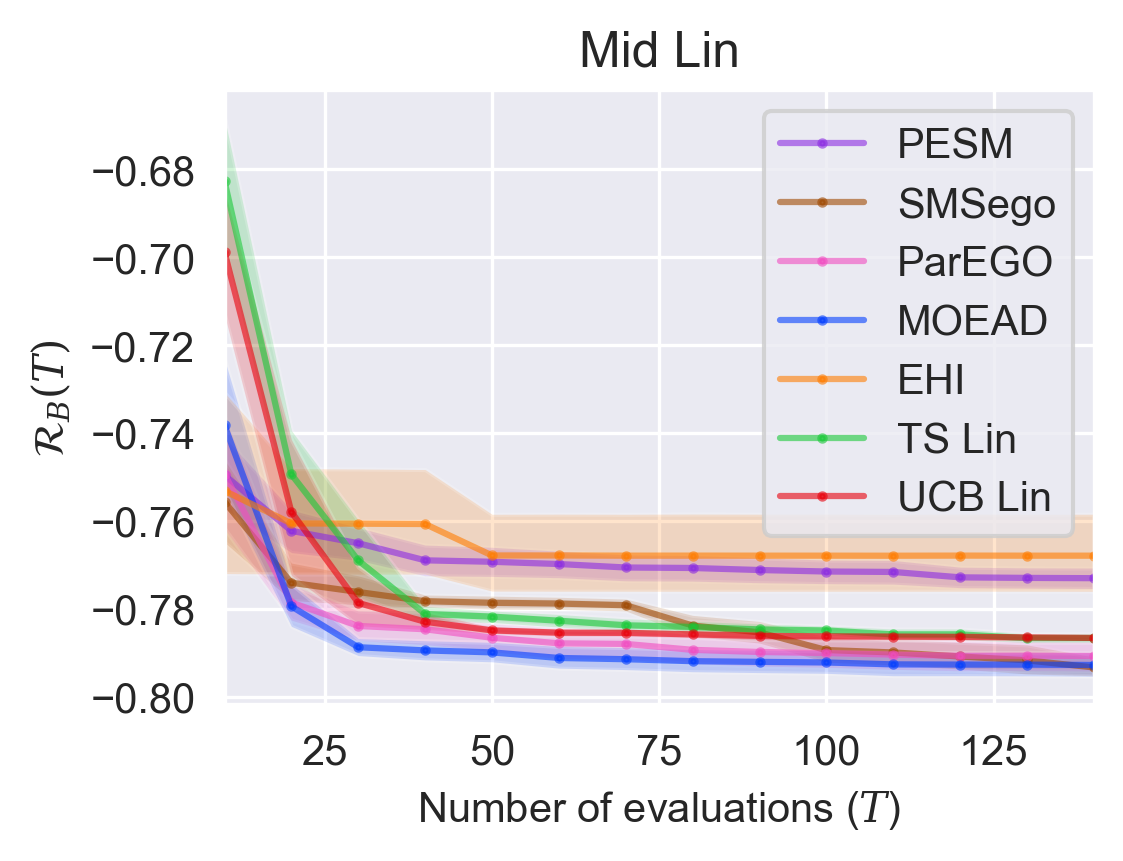}
\end{subfigure}
\begin{subfigure}[t]{0.30\textwidth}
\includegraphics[width=\textwidth]{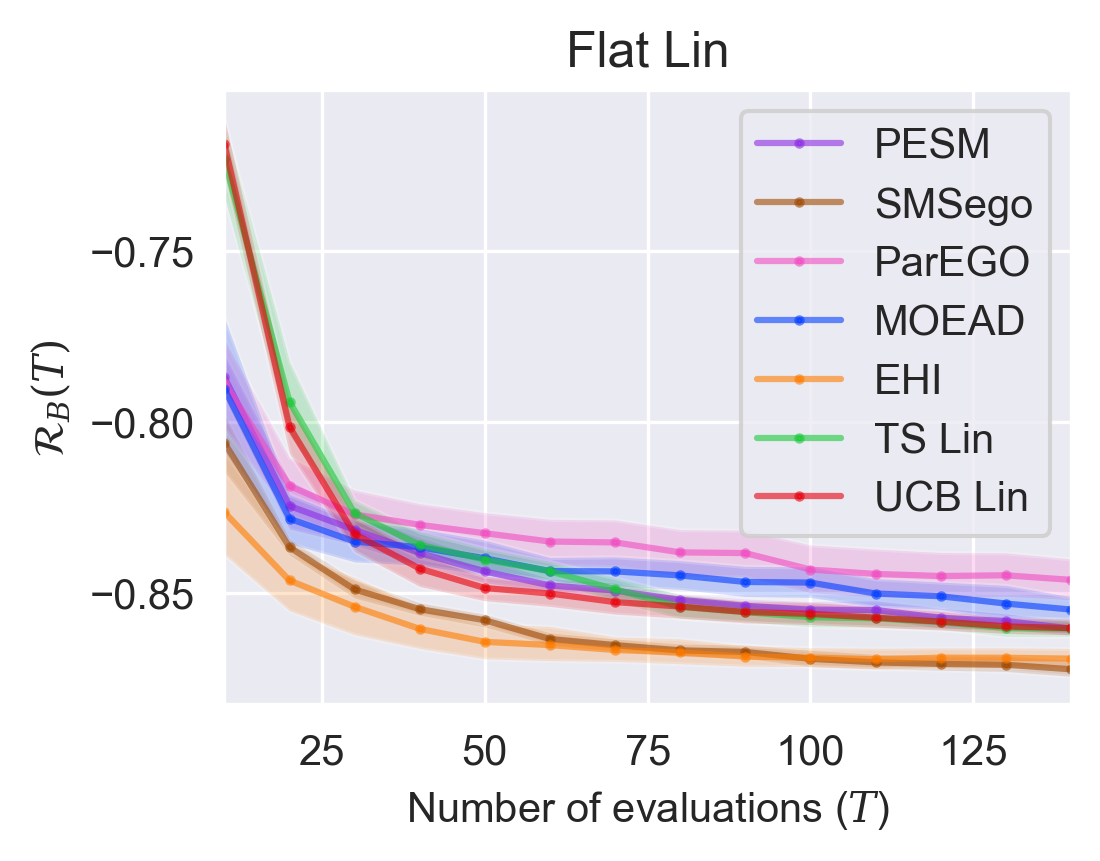}
\end{subfigure}
\caption{Bayes regret plots for the synthetic two-objective function. The mean and the 90\% confidence interval were computed over 10 independent runs. The figure titles denote the sampling region and the scalarization used.}
\label{fig:all_bsr_2d_syn}
\end{figure*}

\begin{figure*}
\centering
\begin{subfigure}[b]{0.30\textwidth}
    \centering
    \includegraphics[width=\textwidth]{./figures/sregret_rand_func_6x6_mid_c.png}
    \subcaption{Synthetic 6x6 function}
\end{subfigure}
\begin{subfigure}[b]{0.30\textwidth}
    \centering
    \includegraphics[width=\textwidth]{./figures/sregret_LSH_glove_mid_c.png}
    \subcaption{LSH Glove}
\end{subfigure}
\begin{subfigure}[b]{0.30\textwidth}
    \centering
    \includegraphics[width=\textwidth]{./figures/sregret_Viola_Jones_mid_c.png}
    \subcaption{Viola Jones}
\end{subfigure}
\begin{subfigure}[b]{0.30\textwidth}
    \centering
    \includegraphics[width=\textwidth]{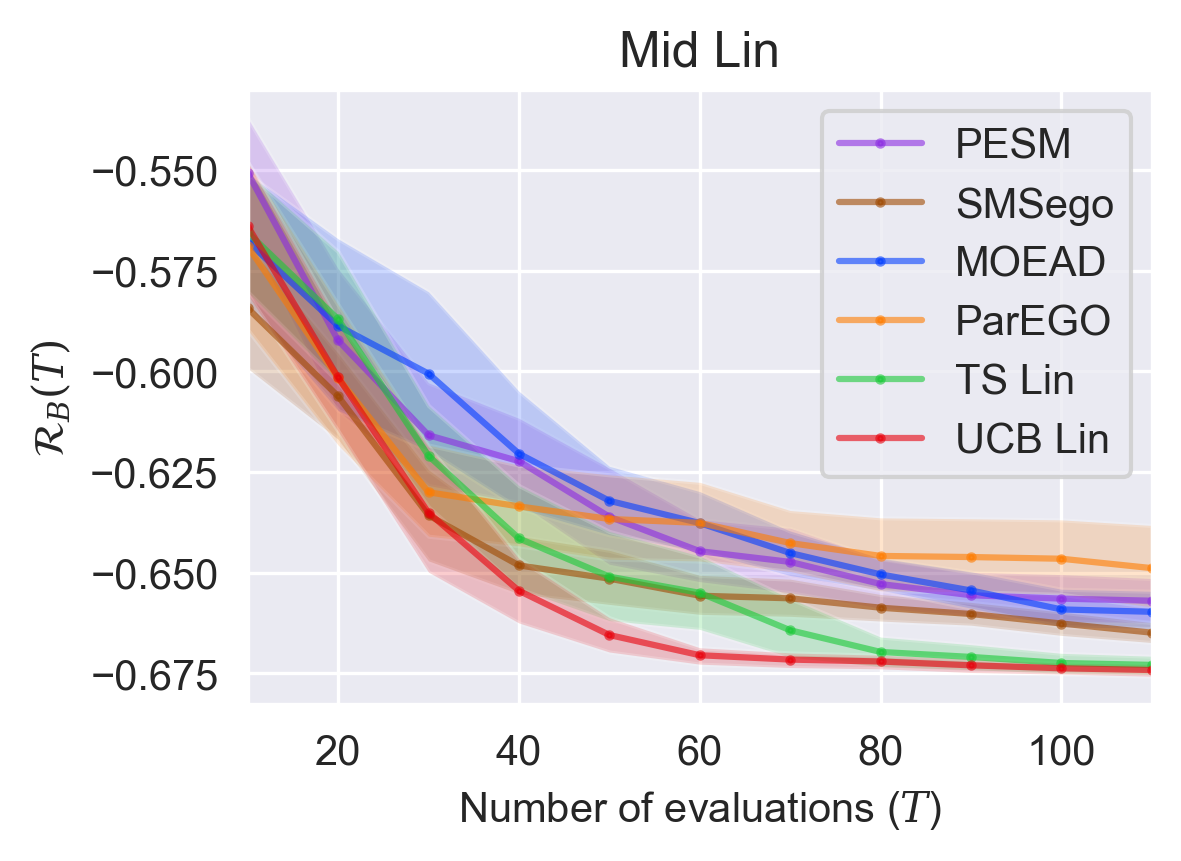}
    \subcaption{Synthetic 6x6 function}
\end{subfigure}
\begin{subfigure}[b]{0.30\textwidth}
    \centering
    \includegraphics[width=\textwidth]{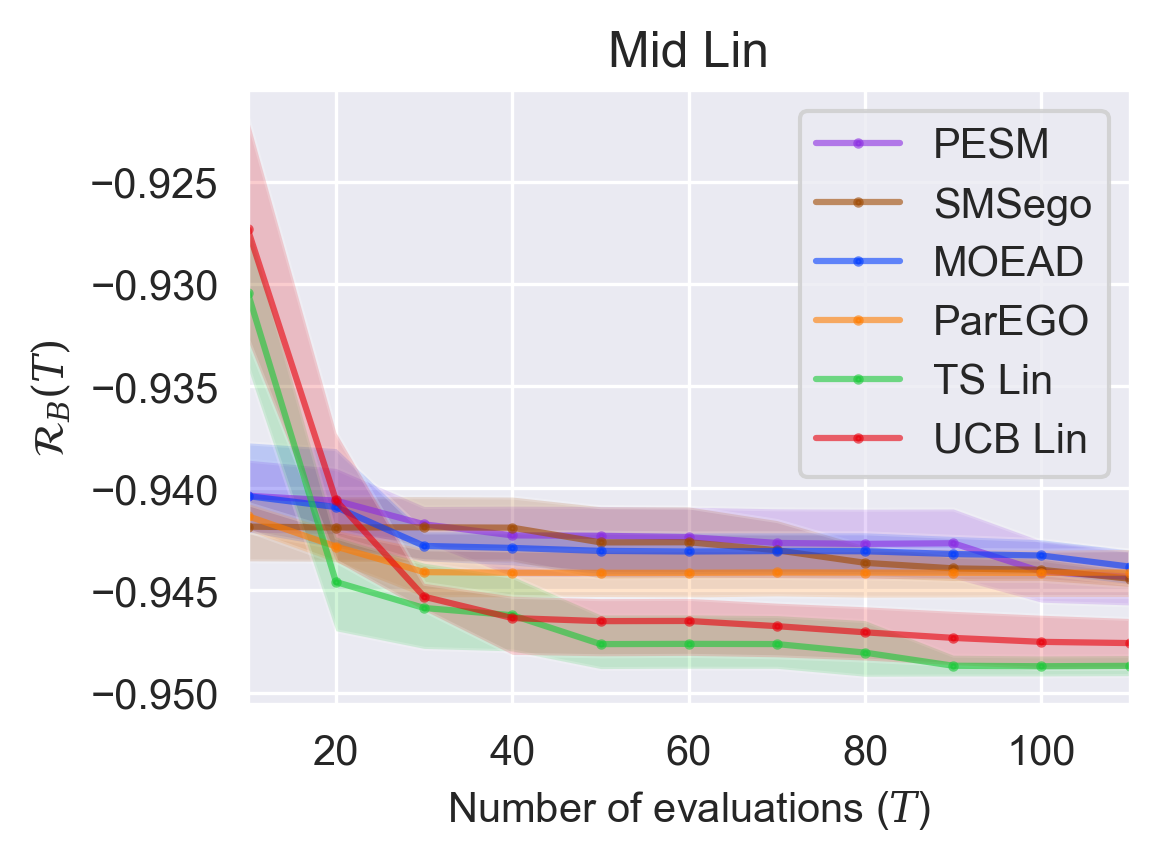}
    \subcaption{LSH Glove}
\end{subfigure}
\begin{subfigure}[b]{0.30\textwidth}
    \centering
    \includegraphics[width=\textwidth]{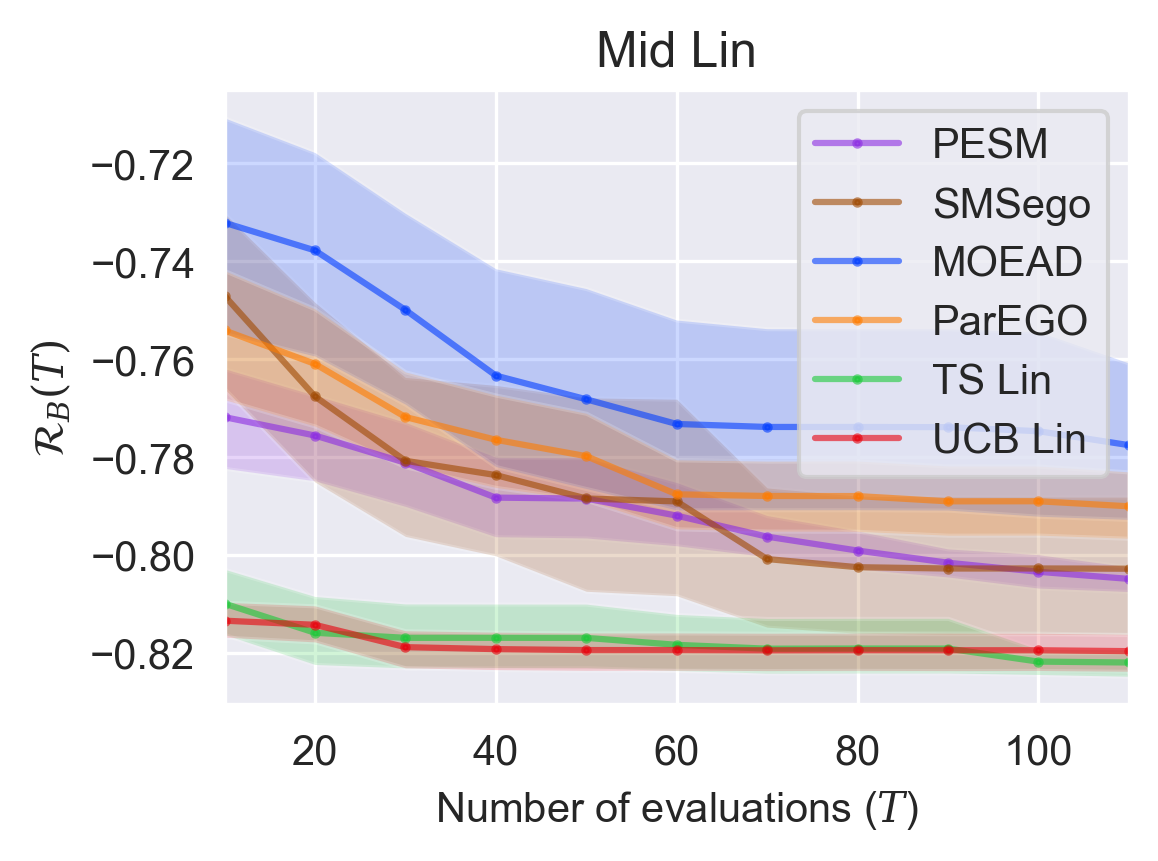}
    \subcaption{Viola Jones}
\end{subfigure}
\caption{Bayes regret plots. The mean and the 90\% confidence interval were computed over 5 runs. The figure titles denote the region sampled and the scalarization used.}
\label{fig:all_bsr_high_dim}
\end{figure*}

\begin{figure*}
\centering
\begin{subfigure}[b]{0.3\textwidth}
\includegraphics[width=\textwidth]{figures/LSH_PESM.png}
\end{subfigure}
\begin{subfigure}[b]{0.3\textwidth}
\includegraphics[width=\textwidth]{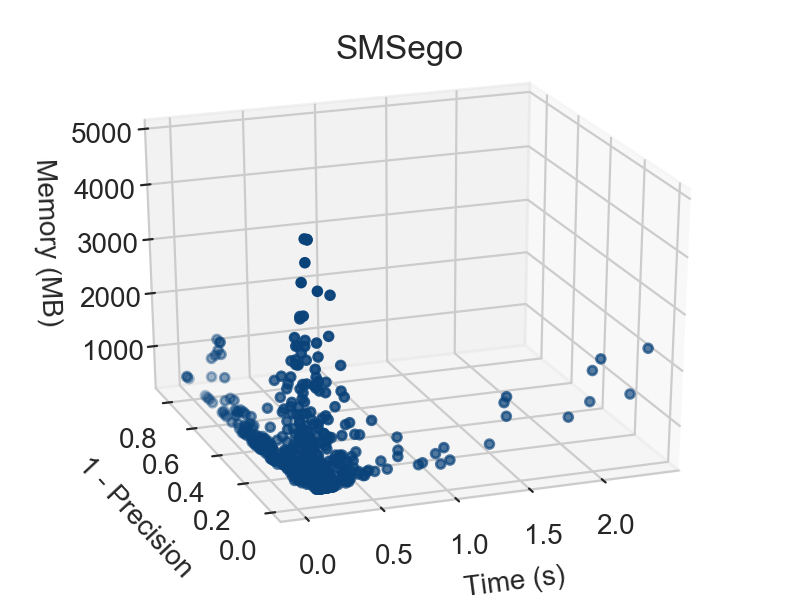}
\end{subfigure}
\begin{subfigure}[b]{0.3\textwidth}
\includegraphics[width=\textwidth]{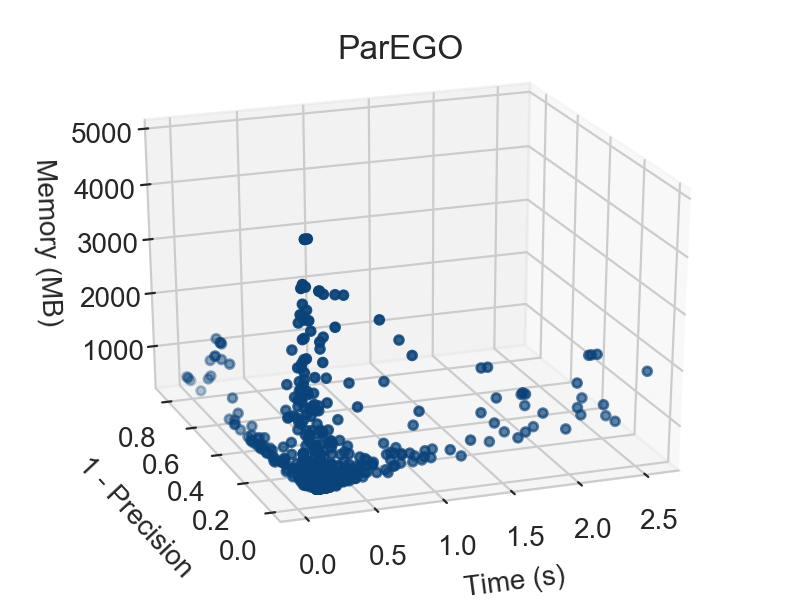}
\end{subfigure}
\begin{subfigure}[b]{0.3\textwidth}
\includegraphics[width=\textwidth]{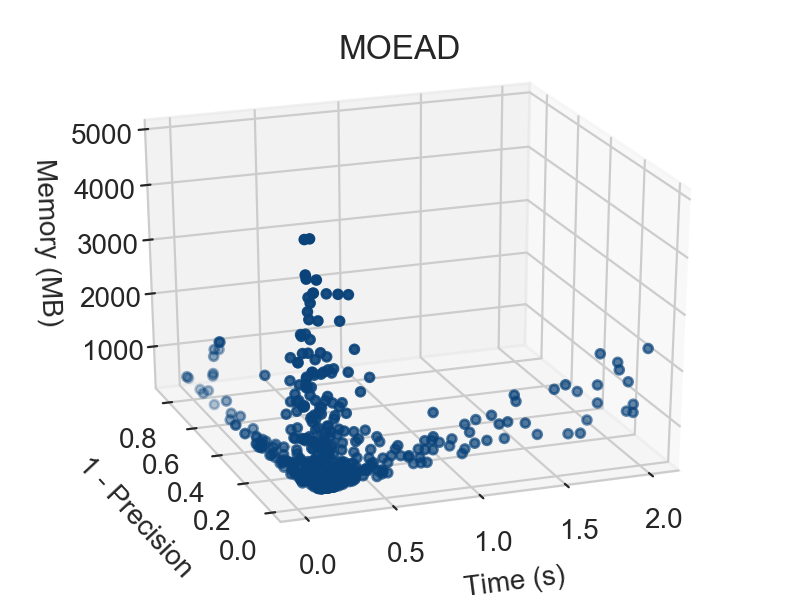}
\end{subfigure}
\begin{subfigure}[b]{0.3\textwidth}
\includegraphics[width=\textwidth]{figures/LSH_ucb_moc_mid.png}
\end{subfigure}
\begin{subfigure}[b]{0.3\textwidth}
\includegraphics[width=\textwidth]{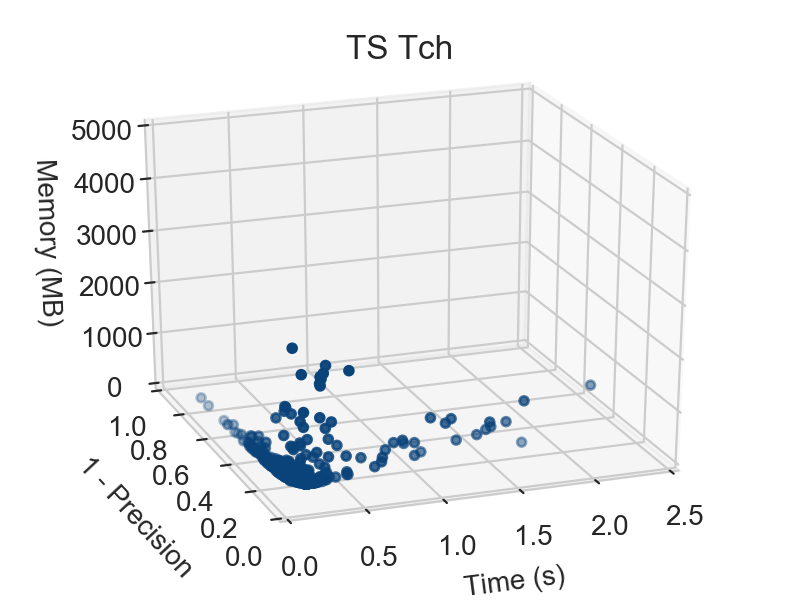}
\end{subfigure}
\begin{subfigure}[b]{0.3\textwidth}
\includegraphics[width=\textwidth]{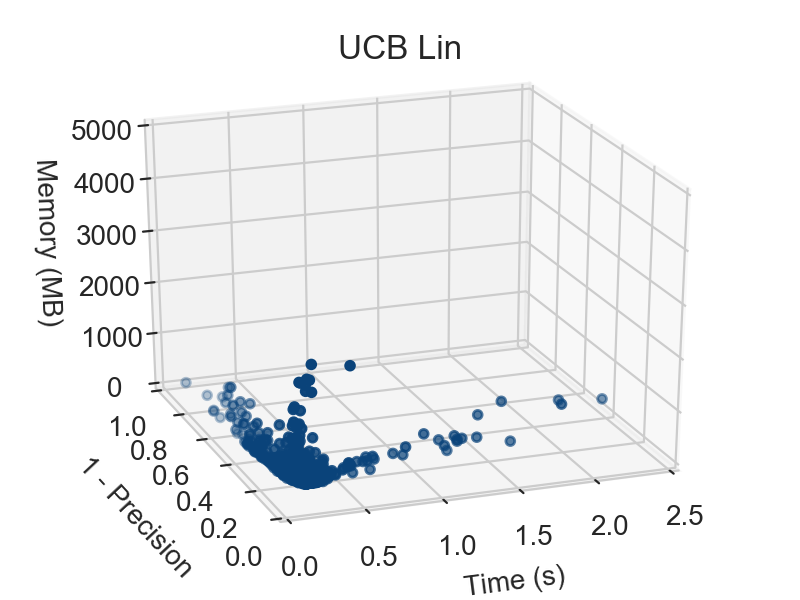}
\end{subfigure}
\begin{subfigure}[b]{0.3\textwidth}
\includegraphics[width=\textwidth]{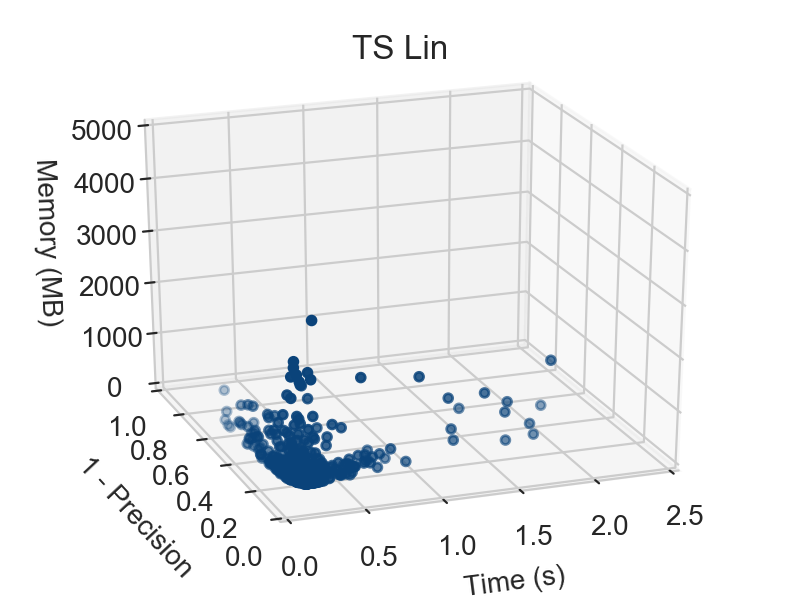}
\end{subfigure}
\caption{Sampled values for the LSH-Glove experiment over 5 independent runs. The figure titles denote the method used.}
\label{fig:all_lsh_samples}
\end{figure*}

Figure~\ref{fig:all_scatter_2d} shows the sampling patterns for all baselines, and all combinations of the sampling region, method and scalarization for our approach. Figure~\ref{fig:all_bsr_2d_syn} show the plots for the Bayes for all sampling regions and scalarizations for the two-objective problem. Figure~\ref{fig:all_bsr_high_dim} shows the Bayes regret for all sampling regions and scalarizations for all the other multi-objective problems. Figure~\ref{fig:all_lsh_samples} shows the sampled objective values for the LSH Glove experiment.

\section{Proofs}
\citet{russo2014learning} introduce a general approach to proving bounds on posterior sampling by decomposing the regret into two sums, one capturing the fact that the UCB upper bounds uniformly with high probability and the other that the UCB is not a loose bound. We begin by making a similar decomposition and bounding each of the other terms. Our proof for TS needs the assumption that the objectives are sampled independently from their respective priors. However, no such assumption is needed for UCB.

Denote by $\Hc_T$ the history until the $T-1$th round $\{(\xv_t, \yv_t, \lambdav_t)\}_{t=1}^{T-1}$. We assume $f_k \sim \Gc\Pc(\zero, \kappa_k)$ have marginal variances upper bounded by 1 for all $\xv \in \Xc$ and $1 \le k \le K$. Let $\xv_t^\star = \argmax_{\xv \in \Xc} \scale_{\lambdav_t}(f(\xv))$. Denote by $U_t(\lambdav, \xv) = \scale_{\lambdav}(\mu^{(t)}(\xv) + \sqrt{\beta_t} \sigma^{(t)}(\xv))$, the UCB as defined in Table~\ref{tab:acq_funcs} where $\mu^{(t)}(\xv),\ \sigma^{(t)}(\xv) \in \Rb^K$ are the posterior means are variances of the $K$ objectives at $\xv$ in step $t$.

We first compute regret bounds for a finite domain $\Xc$, and then use a discretization argument to extend to continuous spaces. We begin by first proving the following decomposition of $\Eb \Rc_C(T)$, and then bound each of the decomposed terms.

\begin{lemma}
\label{lem:decomposition}
For $U_t$ as defined above, the following holds for both UCB and TS.
\begin{multline}
\Eb\Rc_C(T) = \Eb\sbb{\sumT \bb{\maxx \scale_{\lambdav_t}(f(\xv)) - \scale_{\lambdav_t}(f(\xv_t))}}\\
\le \Eb \sbb{\sumT U_t(\lambdav_t, \xv_t) - \scale_{\lambdav_t}(f(\xv_t))} + \\
\Eb \sbb{\sumT \scale_{\lambdav_t}(f(\xv_t^\star)) - U_t(\lambdav_t, \xv_t^\star)}
\end{multline}

where $\xv_t^\star = \argmax_{\xv \in \Xc} \scale_{\lambdav_t}(f(\xv))$.
\end{lemma}

\begin{proof}
For UCB, we use the fact that at each step the next point to evaluate is chosen as $\xv_t = \argmax_{\xv \in \Xc} U_t(\lambdav_t, \xv)$. Thus, conditioned on the history $\Hc_t$, $U_t(\lambdav_t, \xv_t) \ge U_t(\lambdav_t, \xv_t^\star)$. The lemma follows using the tower property of expectation.

Thompson Sampling samples $f_1', \dots, f_k'$ independently from the posterior in each iteration and produces an $\xv_t$ maximizing $\scale_{\lambdav_t}(f(\xv))$. Making use of the independence assumption of the GP priors for TS, we observe that conditioned on the history $\Hc_t$, $\xv_t$ has the same distribution as $\xv_t^\star$, resulting in $\Eb[U_t(\lambdav_t, \xv_t)| \Hc_t] = \Eb[U_t(\lambdav_t, \xv_t^\star)| \Hc_t]$. We use the independence assumption only at this point in the proof and specifically for TS.
\end{proof}

Next we bound both the terms in the decomposition for finite $|\Xc|$, and then use a discretization based argument to prove for continuous sets $\Xc$.

\subsection{Upper Bounds for Finite $|\Xc|$}
\begin{lemma}
\label{lem:bound_1}
For $\beta_t = 2 \ln\bb{\frac{t^2|\Xc|}{\sqrt{2\pi}}}$, and $U_t$ as defined earlier, the following can be bounded as,
\begin{equation}
\Eb \sbb{\sumT \scale_{\lambdav_t}(f(\xv_t^\star)) - U_t(\lambdav_t, \xv_t^\star)} \le \frac{\pi^2}{6} \Eb[L_{\lambdav}] K 
\end{equation}
\end{lemma}

\begin{proof}
We first see that,
\begin{align*}
    & \Eb \sbb{\scale_{\lambdav_t}(f(\xv_t^\star)) - U_t(\lambdav_t, \xv_t^\star)}\\
    & \le \Eb \big(\scale_{\lambdav_t}(f(\xv_t^\star)) - U_t(\lambdav_t, \xv_t^\star)\big)_+\\
    & \le \sum_{\xv \in \Xc} \Eb \big(\scale_{\lambdav_t}(f(\xv)) - U_t(\lambdav_t, \xv)\big)_+
\end{align*}
where $(x)_+$ is defined as $\max(0, x)$. Using Lemma~\ref{lem:bound_lip_mon} and the definition of $U_t$ we can further bound,
\begin{align*}
    & \Eb \big(\scale_{\lambdav_t}(f(\xv)) - U_t(\lambdav_t, \xv)\big)_+ \\
    & \le \Eb[L_{\lambdav}] \sum_{k=1}^K\Eb\Bigsb{\bigb{f(\xv)_k - \mu^{(t)}_k(\xv) - \sqrt{\beta_t} \sigma^{(t)}_k(\xv)}_+}.
\end{align*}

Conditioned on $\Hc_t$, $f(\xv)_k - \mu^{(t)}_k(\xv) - \sqrt{\beta_t} \sigma^{(t)}_k(\xv)$ follows a normal distribution $\Nc\bb{- \sqrt{\beta_t} \sigma^{(t)}_k(\xv),\ {\sigma_k^{(t)}}^2(\xv)}$.

Next we use the fact that for $X\sim \Nc(\mu, \sigma^2)$ and $\mu \le 0$,
\begin{equation}
\label{eqn:normal_upper_bound}
\Eb[X_+] \le \frac{\sigma}{\sqrt{2\pi}} \exp\bb{-\frac{\mu^2}{2\sigma^2}}.
\end{equation}

\begin{align*}
    &\Eb\Bigsb{\bigb{f(\xv)_k - \mu^{(t)}_k(\xv) - \sqrt{\beta_t} \sigma^{(t)}_k(\xv)}_+\ |\ \Hc_t}\\
    & \le \frac{\sigma^{(t)}_k(\xv)}{\sqrt{2\pi}} \exp\bb{-\frac{\beta_t}{2}} \le \frac{1}{t^2|\Xc|}
\end{align*}

Using the tower property of expectation, it follows that,
\begin{equation*}
\Eb \bigsb{\scale_{\lambdav_t}(f(\xv_t^\star)) - U_t(\lambdav_t, \xv_t^\star)} \le \Eb[L_{\lambdav}] \frac{K}{t^2}
\end{equation*}

Summing over $t$, we get
\begin{align*}
& \Eb \sbb{\sumT \scale_{\lambdav_t}(f(\xv_t^\star)) - U_t(\lambdav_t, \xv_t^\star)}\\
& \le \Eb[L_{\lambdav}] K \sum_{t=1}^T\frac{1}{t^2} \le \frac{\pi^2}{6} \Eb[L_{\lambdav}] K 
\end{align*}
completing the proof.
\end{proof}

\begin{lemma}
\label{lem:bound_2}
With the same conditions as in Lemma~\ref{lem:bound_1}, it holds that
\begin{multline}
\bar{L}_{\lambdav}\bb{KT \beta_T \sum_{k=1}^K \frac{\gamma_{Tk}}{\ln(1 + \sigma_k^{-2})}}^{1/2} + \frac{\pi^2}{6}\frac{K\Eb[L_{\lambdav}]}{|\Xc|}
\end{multline}
where $\bar{L}_{\lambdav} = \Eb\sbb{\sqrt{\frac{1}{T} \sum_{t=1}^T L_{\lambda_t}^2}}$.
\end{lemma}

\begin{proof}
Conditioned on $\Cc = (\Hc_{t}, \lambdav_t, \xv_t)$, the following holds using Lemma~\ref{lem:bound_lip_mon},
\begin{align*}
& \Eb \bigsb{U_t(\lambdav_t, \xv_t) - \scale_{\lambdav_t}(f(\xv_t))\ |\ \Cc} \\
& \le \Eb \sbb{L_{\lambdav_t} \sum_{k=1}^K \bb{\mu^{(t)}_k(\xv_t) + \sqrt{\beta_t} \sigma^{(t)}_k(\xv_t) - f(\xv_t)}_+\ \middle|\ \Cc}\\
& \le \Eb  \Bigg[L_{\lambdav_t} \sum_{k=1}^K \bb{\mu^{(t)}_k(\xv_t) + \sqrt{\beta_t} \sigma^{(t)}_k(\xv_t) - f(\xv_t)}\\
&\quad  + L_{\lambdav_t} \sum_{k=1}^K \bb{f(\xv_t) - \mu^{(t)}_k(\xv_t) - \sqrt{\beta_t} 
\sigma^{(t)}_k(\xv_t)}_+ \ \Bigg|\ \Cc\Bigg]\\
& \le \Eb  \Bigg[L_{\lambdav_t} \sum_{k=1}^K \sqrt{\beta_t} \sigma^{(t)}_k(\xv_t) + \frac{KL_{\lambdav_t}}{t^2|\Xc|}\ \Bigg|\ \Cc\Bigg]
\end{align*}
where the last inequality follows from (\ref{eqn:normal_upper_bound}). Using the tower property of expectation, we get
\begin{multline*}
    \Eb \bigsb{U_t(\lambdav_t, \xv_t) - \scale_{\lambdav_t}(f(\xv_t))} \\
    \le \Eb  \Bigg[L_{\lambdav_t} \sqrt{\beta_t} \sum_{k=1}^K \sigma^{(t)}_k(\xv_t) + \frac{KL_{\lambdav_t}}{t^2|\Xc|} \Bigg]
\end{multline*}
Summing over $t$, we get,
\begin{align*}
    & \Eb \sbb{\sum_{t=1}^T U_t(\lambdav_t, \xv_t) - \scale_{\lambdav_t}(f(\xv_t))} \\
    & \le \Eb  \sbb{\sum_{k=1}^K \sum_{t=1}^T L_{\lambdav_t} \sqrt{\beta_t}\sigma^{(t)}_k(\xv_t) + \sum_{t=1}^T \frac{KL_{\lambdav_t}}{t^2|\Xc|}}\\
    & \le \Eb  \sbb{\sum_{k=1}^K \sum_{t=1}^T L_{\lambdav_t} \sqrt{\beta_t}\sigma^{(t)}_k(\xv_t)} + \frac{\pi^2}{6}\frac{K\Eb[L_{\lambdav}]}{|\Xc|}\\
    & \le \Eb  \Bigg[\bb{K\beta_T\sum_{t=1}^T L^2_{\lambdav_t}}^{1/2} \\
    & \quad \bb{\sum_{k=1}^K \sum_{t=1}^T{\sigma^{(t)}_k}^2(\xv_t)}^{1/2}\Bigg] + \frac{\pi^2}{6}\frac{K\Eb[L_{\lambdav}]}{|\Xc|}\\
    & \le \Eb  \Bigg[\bb{K\beta_T\sum_{t=1}^T L^2_{\lambdav_t}}^{1/2} \quad \bb{\sum_{k=1}^K \frac{\gamma_{Tk}}{\ln(1 + \sigma_k^{-2})}}^{1/2}\Bigg] \\
    & \quad + \frac{\pi^2}{6}\frac{K\Eb[L_{\lambdav}]}{|\Xc|}
\end{align*}
where the second last step follows using Cauchy-Schwartz inequality, and the last step used the upper bound in terms of the MIG as shown in \citet{srinivas2010gaussian}. Substituting $\bar{L}_{\lambdav}$ gives us the desired result.
\end{proof}

\begin{proposition}
The cumulative regret incurred in Algorithm~\ref{alg:moors} for both UCB and TS is upper bounded as,
\begin{equation}
    \bar{L}_{\lambdav}\bb{KT \beta_T \sum_{k=1}^K \frac{\gamma_{Tk}}{\ln(1 + \sigma_k^{-2})}}^{1/2} + \frac{\pi^2}{3}K\Eb[L_{\lambdav}]
\end{equation}
\end{proposition}
\begin{proof}
The proof follows directly using Lemmas~\ref{lem:decomposition},~\ref{lem:bound_1}, and~\ref{lem:bound_2}.
\end{proof}

\subsection{Extending to continuous $\Xc$}
We begin with the following result due to \citet{ghosal2006posterior}. For any differentiable stationary kernel $\kappa$ with $4$th order derivatives and $f\sim \GP(\zero, \kappa)$, we have the following bound holds for some $a, b > 0$ such that for all $J > 0$, and for all $i \in \{1, \dots, d\}$,
\begin{equation}
    \Pb \bb{\sup_x \left|\parder{f(x)}{x_i}\right| > J} \le ae^{-(J/b)^2}.
\end{equation}

Consider a continuous set $\Xc$ where $\Xc \subset \Rb^d$. For the sake of analysis, at each time step $t$ we construct a finite discretization $\Xc_t$ of $\Xc$. $\Xc_t$ is constructed using a grid of uniformly spaced points with a distance of $\tau_j^{-1}$ between adjacent points in each coordinate. Therefore $|\Xc_t| = \tau_j^d$. Let $[\xv]_t$ denote the point closest to $\xv$ in $\Xc_t$. Let $M = \sup_{i\in \{1,\dots, d\}}\sup_{\xv\in \Xc} \left|\parder{f(\xv)}{x_i}\right|$. $\Eb\big[|f(\xv) - f([\xv]_t)|\big]$ can be bounded as
\begin{align*}
    & \Eb\big[|f(\xv) - f([\xv]_t)|\big]\\
    & \le \frac{d}{\tau_t} \Eb[M] \le \frac{d}{\tau_t} \int_{0}^\infty \Pb(M\ge t) dt\\
    & \le \frac{d}{\tau_t} \int_{0}^\infty ae^{-(t/b)^2} dt = \frac{dab\sqrt{\pi}}{2\tau_t}
\end{align*}
Let $A = \sup_{k=1}^K a_k,\ B = \sup_{k=1}^K b_k$ where $a_k,\ b_k$ correspond to the above constants for the $k$th objective. It follows that,
\begin{equation}
    \Eb\bigsb{|\scale_{\lambdav}(f(\xv)) - \scale_{\lambdav}(f([\xv]_t))|} \le K\Eb[L_{\lambdav}]\frac{dAB\sqrt{\pi}}{2\tau_t}
\end{equation}

We choose $\tau_t = t^2dAB\sqrt{\pi}/2$ which gives us
\begin{equation}
    \Eb\bigsb{|\scale_{\lambdav}(f(\xv)) - \scale_{\lambdav}(f([\xv]_t))|} \le K\Eb[L_{\lambdav}]\frac{1}{t^2}
    \label{eqn:approx_x_t}
\end{equation}

Having bounded the approximation errors due to the discretization, we are in a position to use the framework developed for the finite case. We begin with a similar decomposition as Lemma~\ref{lem:decomposition}, which includes the approximation factors \citep{kandasamy2018parallelised}. For the continuous case, our treatment differs slightly for TS and UCB. We first look at the decomposition for UCB,

\begin{lemma}
For the same conditions as in Lemma~\ref{lem:decomposition}, and $[\xv]_t$ as defined above, we have the following decomposition for TS,

\textbf{TS:}
\begin{multline}
    \Eb\Rc_C(T) \le \underbrace{\Eb \sbb{\sumT \scale_{\lambdav_t}(f([\xv_t]_t)) - \scale_{\lambdav_t}(f(\xv_t))}}_{A1} + \\
    \underbrace{\Eb \sbb{\sumT U_t(\lambdav_t, [\xv_t]_t) - \scale_{\lambdav_t}(f([\xv_t]_t))}}_{A2} + \\
    \underbrace{\Eb \sbb{\sumT \scale_{\lambdav_t}(f([\xv_t^\star]_t)) - U_t(\lambdav_t, [\xv_t^\star]_t)}}_{A3} + \\
    \underbrace{\Eb \sbb{\sumT \scale_{\lambdav_t}(f(\xv_t^\star)) - \scale_{\lambdav_t}(f([\xv_t^\star]_t))}}_{A4}
\end{multline}
\end{lemma}

The proof is on the same lines as Lemma~\ref{lem:decomposition} using the fact that $\xv_t^\star$ and $\xv_t$ have the same distribution, when conditioned on the $\Hc_t$.

We now bound each of the individual terms. $A1$ and $A4$ can be bounded by $C_1 K\Eb[L_{\lambdav}]$ using (\ref{eqn:approx_x_t}), for some global constant $C_1$. Let $\beta_t = \ln\bb{\frac{t^2 |\Xc_t|}{\sqrt{2\pi}}}$. $A2 + A3$ can be bounded in the same way as Lemma~\ref{lem:bound_1} and~\ref{lem:bound_2} by considering the discretized set $\Xc_t$ at each time step $t$ instead of $\Xc$. 

For UCB, we have the following decomposition,

\begin{lemma}
For the same conditions as in Lemma~\ref{lem:decomposition}, and $[\xv]_t$ as defined above, we have the following decomposition for UCB,

\textbf{UCB:}
\begin{multline}
    \Eb\Rc_C(T) \le \underbrace{\Eb \sbb{\sumT U_t(\lambdav_t, \xv_t) - \scale_{\lambdav_t}(f(\xv_t))}}_{B1} + \\
    \underbrace{\Eb \sbb{\sumT \scale_{\lambdav_t}(f([\xv_t^\star]_t)) - U_t(\lambdav_t, [\xv_t^\star]_t)}}_{B2} + \\
    \underbrace{\Eb \sbb{\sumT \scale_{\lambdav_t}(f(\xv_t^\star)) - \scale_{\lambdav_t}(f([\xv_t^\star]_t))}}_{B3}
\end{multline}
\end{lemma}

The proof follows in a similar way as Lemma~\ref{lem:decomposition} using the fact that $U_t(\lambdav_t, \xv_t) \ge U_t(\lambdav_t, [\xv_t^\star]_t)$.

We now bound each of the decomposed terms by considering the discretized $\Xc_t$ in each step, and the corresponding $\beta_t$. $B3$ can be bounded in the same way as $A4$ using (\ref{eqn:approx_x_t}). $B1$ can be bounded using Lemma~\ref{lem:bound_2}. $B2$ can be bounded in the same way as Lemma~\ref{lem:bound_1}.

This leads us to the following theorem.

\begin{theorem}
\label{thm:final}
For $\beta_t = \ln\bb{\frac{t^2 |\Xc_t|}{\sqrt{2\pi}}}$, where $|\Xc_t| = \bb{\frac{t^2dAB \sqrt{\pi}}{2}}^d$, for some global constants $C_1, C_2 > 0$, the following holds for both UCB and TS,
\begin{multline}
    \Eb\Rc_C(T) \le C_1K\Eb[L_{\lambdav}] + \\
    C_2 \bar{L}_{\lambdav}\bb{KT (d\ln T + d\ln d) \sum_{k=1}^K \frac{\gamma_{Tk}}{\ln(1+\sigma_k^{-2})}}^{1/2}
\end{multline}
\end{theorem}

\subsection{Upper bound on Bayes regret}
\label{sec:br_upper_bound}
At a high level, optimizing the cumulative regret optimizes the pointwise regret for the observed $\lambdav_t$. However, it generalizes to the unseen $\lambdav$ in $\Rc_B(T)$ that are \emph{close} to the sampled $\lambdav_t$. This requires us to define a notion of closeness or a metric on $\Lambda$. We assume that $\Lambda$ is a bounded subset of a $\Rb^D$. We make the assumption that $\scale_{\lambdav}(\yv)$ is $J$-Lipschitz in $\lambdav$ for all $\yv \in \Rb^K$, that is,
\begin{equation}
    |\scale_{\lambdav_1}(\yv) - \scale_{\lambdav_2}(\yv)| \le J\,\|\lambdav_1 - \lambdav_2\|_1.
\end{equation}

Conditioned on the history $\Hc_T$, consider the Wasserstein or Earth Movers distance \citep{monge1781memoire, villani2008optimal} $W_1(p, \widehat{p})$ between the sampling distribution $p(\lambdav)$ defined on $\Lambda$, and the empirical distribution $\widehat{p}$ corresponding to the samples $\{\lambdav_t\}_{t=1}^T$,
\begin{multline}
    W_1(p, \widehat{p}) =\\
    \inf_q\big\{\Eb_q\| X - Y \|_1,\ q(X) = p,\  q(Y) = \widehat{p}\big\},
\end{multline}
where $q$ is a joint distribution on the RVs $X, Y$, with marginal distributions equal to $p$ and $\widehat{p}$. We then have,
\begin{align*}
    & \frac{1}{T}\sum_{t=1}^T \scale_{\lambdav_t}(f(\xv_t)) - \Eb\sbb{\max_{x\in\Xv} \scale_{\lambdav}(f(\xv))}\\
    & \le \frac{1}{T}\sum_{t=1}^T \max_{\xv \in \Xv} \scale_{\lambdav_t}(f(\xv)) - \Eb\sbb{\max_{\xv \in\Xv} \scale_{\lambdav}(f(\xv))}\\
    & \le \Eb_{q(X, Y)}\sbb{\max_{\xv \in \Xv} \scale_{Y}(f(\xv)) - \max_{\xv \in \Xv} \scale_{X}(f(\xv))}\\
    & \le \Eb_{q(X, Y)}\big\{J \|X-Y\|_1\big\}
\end{align*}
Taking the infimum w.r.t. $q$, and the expectation w.r.t. the history $\Hc_t$, we get,
\begin{multline}
    \Eb\sbb{\frac{1}{T}\sum_{t=1}^T \scale_{\lambdav_t}(f(\xv_t))} - \Eb\sbb{\max_{x\in\Xv} \scale_{\lambdav}(f(\xv))} \\
    \le J\,\Eb W_1(p, \widehat{p})
\end{multline}
Using the fact that $\Eb\sbb{\maxx \scale_{\lambdav}(f(\xv))} = \Eb\sbb{\maxx \scale_{\lambdav_t}(f(\xv))}$, we get,
\begin{equation}
    \Eb\Rc_B(T) \le \underbrace{\frac{1}{T} \Eb\Rc_C(T)}_{\mathbf{I}} + J\,\underbrace{\Eb W_1(p, \widehat{p})}_{\mathbf{II}}.
\end{equation}
As $T \to \infty$, $\mathbf{I}$ converges to zero at a rate of $\Oc^*(T^{-1/2})$\footnote{$\Oc^*$ ignores logarithmic factors.} as given by Theorem~\ref{thm:final}. Term $\mathbf{II}$ converges to zero at a rate of $\Oc^*(T^{-1/D})$ when $D \ge 2$, under mild regulatory conditions as shown by \citet{canas2012learning}.

\subsection{Auxilliary Results}
\begin{lemma}
\label{lem:bound_lip_mon}
Suppose $s: \Rb^D \to \Rb$ is $L$-Lipschitz in the $\ell_1$-norm, and monotonically increasing in all coordinates. Then it holds that,
\begin{equation}
    \bigb{s(\xv) - s(\yv)}_+ \le L\sum_{d=1}^D (\xv_d - \yv_d)_+
\end{equation}
where $(x)_+$ is defined as $\max(0, x)$.
\end{lemma}

\begin{proof}
We first note that when $s(\xv) \le s(\yv)$, it holds trivially. Now we assume $s(\xv) > s(\yv)$. Let $U_d,\ 0\le d \le D$ be defined as
\begin{equation*}
    U_d = \begin{cases}
        s(\xv),\quad\text{if } d = 0\\
        s(\yv),\quad\text{if } d = D\\
        s(\yv_1, \dots, \yv_{d}, \xv_{d+1}, \dots, \xv_D),\quad \text{otherwise}
    \end{cases}
\end{equation*}

Then,
\begin{align*}
    0\le s(\xv) - s(\yv) = \sum_{d=0}^{D-1} U_d - U_{d+1}.
\end{align*}
Using the facts that $U_d,\ U_{d+1}$ differ only in the $d+1$ component, and that $s$ is increasing in all the components, we get
\begin{align*}
    & 0\le s(\xv) - s(\yv)\\
    & \le \sum_{d=0}^{D-1} |U_d - U_{d+1}| \Ib\bigb{\xv_{d+1} - \yv_{d+1} \ge 0}\\
    & \le \sum_{d=0}^{D-1} L|\xv_{d+1} - \yv_{d+1}| \Ib\bigb{\xv_{d+1} - \yv_{d+1} \ge 0}\\
    & = L \sum_{d=1}^{D} (\xv_{d} - \yv_{d})_+
\end{align*}
concluding the proof.
\end{proof}

\section{Implementation Details}
The two-objective experiment was repeated 10 times with 150 iterations per run. All other experiments were repeated 5 time with 120 iterations per run. MOEA/D-EGO supports batch evaluation of points in every iteration. However, for fair comparison with the other methods, we set the batch size to 1.

\textbf{Domain space:} For all our experiments we map the input domain appropriately such that $\Xc = [0, 1]^d$.

\textbf{Initial evaluations:} Similar to \citet{kandasamy2015high}, we randomly choose $n_{\mathrm{init}}$ initial points. We then evaluate the MO function at the initial points before using our optimization strategy.

\textbf{Hyper-parameter estimation:} To estimate the GP hyper-parameters, the GP is fitted to the observed data every 10 evaluations. We use the squared exponential kernel for all our experiments. We have a separate bandwidth parameter for each dimension of the input domain. The bandwidth, scale and noise variance are estimated by maximizing the marginal likelihood \citep{rasmussen2006gaussian}. We set the mean of the GP as the median of the observations.

\textbf{UCB parameter $\beta_t$:} As discussed in \citet{kandasamy2015high}, $\beta_t$ as suggested in \cite{srinivas2010gaussian} is too conservative in practice, and with unknown constants. Following the recommendation in \cite{kandasamy2015high}, we use $\beta_t = 0.125 \log(2t + 1)$ for all our experiments.

\textbf{Optimizing the acquisition function} We use the DiRect algorithm \citep{jones1993lipschitzian} for optimizing the acquisition function in each iteration.

\end{document}